\documentclass{article}

\usepackage{microtype}
\usepackage{graphicx}
\usepackage{booktabs} 
\usepackage{dsfont}

\usepackage{hyperref}

\usepackage[accepted]{icml2024}

\usepackage{amsmath}
\usepackage{amssymb}
\usepackage{mathtools}
\usepackage{amsthm}

\usepackage[capitalize,noabbrev]{cleveref}

\usepackage{hyperref}       
\usepackage{url}            
\usepackage{booktabs}       
\usepackage{amsfonts}       
\usepackage{nicefrac}       
\usepackage{microtype}      
\usepackage{xcolor}         
\usepackage{graphicx}
\usepackage{caption}
\usepackage{bbm}
\usepackage{subcaption}
\usepackage{multirow}
\usepackage{float}
\usepackage{amsmath}
\usepackage{amssymb}
\usepackage{stmaryrd}

\DeclareMathOperator{\E}{\mathbb{E}}

\theoremstyle{plain}

\theoremstyle{definition}

\theoremstyle{remark}

\usepackage[textsize=tiny]{todonotes}

\icmltitlerunning{Confidence Aware Inverse Constrained Reinforcement Learning}

\begin{document}

\twocolumn[
\icmltitle{Confidence Aware Inverse Constrained Reinforcement Learning}



\icmlsetsymbol{equal}{*}

\begin{icmlauthorlist}
\icmlauthor{Sriram Ganapathi Subramanian}{yyy}
\icmlauthor{Guiliang Liu}{comp}
\icmlauthor{Mohammed Elmahgiubi}{sch}
\icmlauthor{Kasra Rezaee}{sch}
\icmlauthor{Pascal Poupart}{sch2,yyy}
\end{icmlauthorlist}

\icmlaffiliation{yyy}{Vector Institute for Artificial Intelligence, Toronto, Canada}
\icmlaffiliation{comp}{School of Data Science, The Chinese University of Hong Kong, Shenzhen,
Guangdong, 518172, P.R. China}
\icmlaffiliation{sch}{Huawei Technologies Canada}
\icmlaffiliation{sch2}{Cheriton School of Computer Science, University of Waterloo, Canada}

\icmlcorrespondingauthor{Sriram Ganapathi Subramanian}{sriram.subramanian@vectorinstitute.ai}

\icmlkeywords{Machine Learning, ICML}

\vskip 0.3in
]



\printAffiliationsAndNotice{}  

\begin{abstract}
In coming up with solutions to real-world problems, humans implicitly adhere to constraints that are too numerous and complex to be specified completely. However, reinforcement learning (RL) agents need these constraints to learn the correct optimal policy in these settings. The field of Inverse Constraint Reinforcement Learning (ICRL) deals with this problem and provides algorithms that aim to estimate the constraints from expert demonstrations collected offline. Practitioners prefer to know a measure of confidence in the estimated constraints, before deciding to use these constraints, which allows them to only use the constraints that satisfy a desired level of confidence.  However, prior works do not allow users to provide the desired level of confidence for the inferred constraints. This work provides a principled ICRL method that can take a confidence level with a set of expert demonstrations and outputs a constraint that is at least as constraining as the true underlying constraint with the desired level of confidence. Further, unlike previous methods, this method allows a user to know if the number of expert trajectories is insufficient to learn a constraint with a desired level of confidence, and therefore collect more expert trajectories as required to simultaneously learn constraints with the desired level of confidence and a policy that achieves the desired level of performance. 
\end{abstract}

\section{Introduction}
Reinforcement learning (RL) \cite{sutton2018reinforcement} has seen large successes in a variety of computer games like Atari \cite{mnih2015human} and StarCraft \cite{vinyals2019grandmaster}, and more recently in real-world environments like recommender systems \cite{afsar2022reinforcement} and robotics \cite{zhao2020sim}. Traditionally in RL, agents are allowed to explore the entire state and action space to learn the optimal policy. However, in many real-world environments, considerations of safety and feasibility prevent an exploring agent from visiting all states and actions~\cite{dulac2021challenges, ray2019benchmarking}. This observation led to the development of constrained reinforcement learning (CRL) \cite{liu2021policy} where the policy is constrained to remain within the bounds of a set of constraint functions that limit the states and/or actions that an agent can explore in the environment. However, it is hard for human designers to fully specify all constraints in complex real-world environments. Alternatively, it is possible to obtain expert demonstrations. For example, in autonomous driving, it is far easier to obtain demonstrations from expert human drivers that illustrate the optimal behavior in different road and weather conditions. Using these demonstrations, the agent will need to infer the underlying constraints that the expert follows to ensure smoothness, comfort and safety.

\begin{figure*}
\centering
\begin{subfigure}{.5\textwidth}
  \centering
\includegraphics[width=6cm]{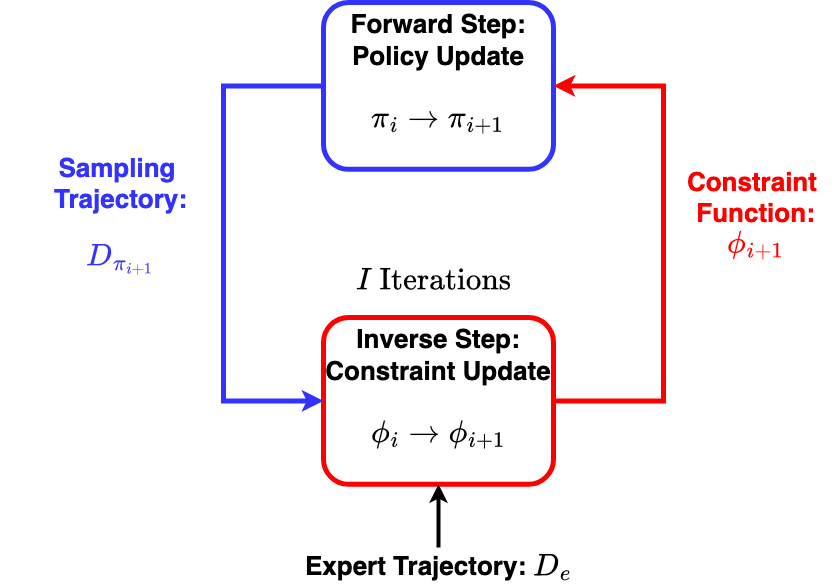}
  \caption{Inverse Constrained RL}
  \label{fig:ICRL}
\end{subfigure}%
\begin{subfigure}{.5\textwidth}
  \centering
  \includegraphics[width=6.5cm]{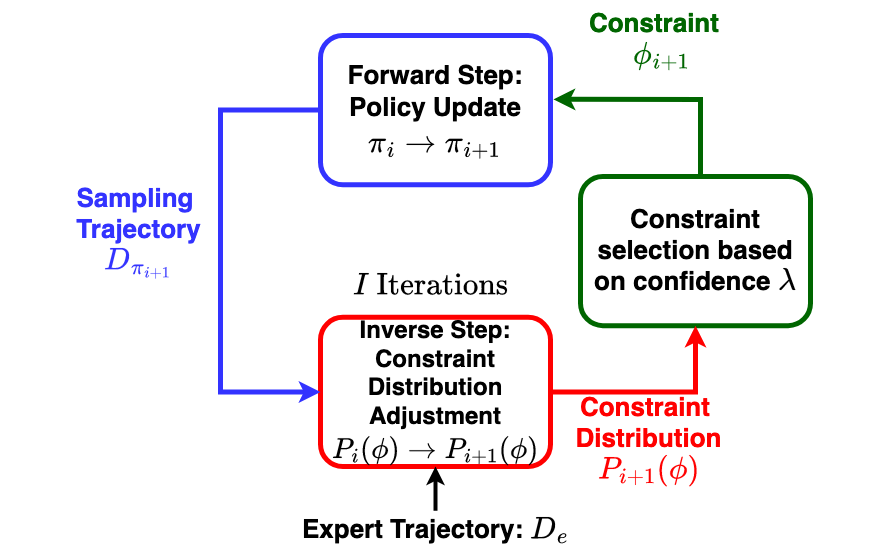}
  \caption{Confidence Aware - Inverse Constrained RL}
  \label{fig:CA-ICRL}
\end{subfigure}
\caption{A figure showing the architectures of ICRL and CA-ICRL}
\label{fig:test}
\end{figure*}


The field of Inverse Constrained Reinforcement Learning (ICRL) \cite{malik2021inverse} aims to solve the problem of learning the underlying constraints from expert demonstrations. This is a new field of research where the algorithms have a structure of alternating between updating a constraint function and learning a policy that satisfies its current constraint estimate (see Figure~\ref{fig:ICRL}). While several recent works introduce strong algorithms for ICRL \cite{malik2021inverse, liu2023benchmarking, gaurav2023learning,qiao2023multimodal,Liu2023OnlineICRL,liu2024meta,xu2024uncertainty}, these methods estimate only the constraint function, but do not provide an estimate of the confidence in the learned constraint. Intuitively, agents will be more confident in the learned constraint if there is a greater number of expert trajectories demonstrating behavior that follows such a constraint. In this paper, we provide an algorithm, called \emph{confidence-aware ICRL (CA-ICRL)} that uses a confidence estimate along with learning the constraint itself. The algorithm will accept a desired confidence and learn a constraint that is at-least as constraining as the ground truth constraint with the provided confidence. Humans and/or agents can choose to learn appropriate constraints based on their risk profiles. Also, this allows for the identification of high-confidence constraints for learning optimal policies. Additionally, CA-ICRL can also be used to decide if more expert trajectories need to be collected based on the desired levels of confidence in constraint and the desired performance. The architecture of this approach is given in Figure~\ref{fig:CA-ICRL}, where the algorithm iterates between updating a forward policy and updating its estimate of constraints like the vanilla ICRL approach, and in addition using the confidence to select the constraint. An elaborate discussion of prior works in ICRL is available in Appendix~\ref{appendix:relatedwork}. 



\section{Background}

We describe the fundamentals of constrained reinforcement learning (CRL) and inverse constrained reinforcement learning (ICRL) in this section. 

\subsection{Constrained Reinforcement Learning} 

The CRL approach solves the constrained Markov decision process (CMDP) \cite{altman1999constrained}, which aims to learn an optimal policy that maximizes discounted cumulative rewards while ensuring that the agent satisfies a set of safety constraints. The CMDP can be seen as an extension of the standard Markov decision process (MDP)  $\langle S, A, P_T, P_R, \mu, \gamma \rangle$ \cite{sutton2018reinforcement}, with a constraint set $\mathcal{C}$. Here $S$ represents the state space, $A$ represents the action space, $P_T(s'|s, a)$ indicates the probability of transitioning to state $s'$ after executing action $a$ in state $s$, $P_R(r|s,a)$ indicates the probability of earning reward $r$ in state $s$ when executing action $a$, $\mu(s)$ is the probabilty that the initial state is $s$, and $\gamma\in[0,1)$ represents the discount factor. We will also use $\bar{r}(s,a)=E_{P_R}[r|s,a]$ to denote the expected reward of a state-action pair.  Actions are chosen according to a stochastic policy $\pi(a|s)\in[0,1]$ that indicates the probability with which action $a$ is chosen in state $s$. The field of RL tries to find an optimal policy $\pi^*$ that maximizes expected discounted cumulative rewards. 
\begin{equation}
\pi^* = \arg \max_\pi E_{P_\pi}[\bar{r}(\tau)]
\end{equation}
Here $\tau=(s_0,a_0,s_1,a_1,s_2,a_2,...)$ denotes a trajectory of state-action pairs, $P_\pi(\tau)=\mu(s_0)\prod_{t=0}^\infty \pi(a_t|s_t) P_T(s_{t+1}|s_t,a_t)$ denotes the probability that policy $\pi$ will yield trajectory $\tau$ and $\bar{r}(\tau)=\sum_{t=0}^\infty \bar{r}(s_t, a_t)$ is the expected cumulative reward of trajectory $\tau$. In the maximum entropy framework, an additional entropy term $H(P_\pi)=-\int_{\tau} P_\pi(\tau) \log P_\pi(\tau)$ is added to the expected rewards.  This term encourages stochasticity in action choices which provides a form of regularization and some degree of exploration. 
\begin{equation}
\pi^*_{maxent} = \arg \max_\pi \E_{P_\pi}[\bar{r}(\tau)] + \beta H(P_\pi)
\end{equation}
In many application domains, policies must satisfy constraints that can be incorporated in a Constrained Markov decision process CMDP represented as $\langle S, A, P_T, P_R, \mu, \gamma, \mathcal{C} \rangle$, which extends the MDP with a constraint set $\mathcal{C} = \{ (P_{C,i}, \epsilon_i) \}_{i=1}^m$, where $P_{C,i}(c|s,a)$ denotes the probability that cost $c$ is incurred in state $s$ and action $a$ in accordance with the $i^{th}$ constraint, and $\epsilon_i$ is an upper bound on expected cumulative costs in the $i^{th}$ constraint.  It is often convenient to refer to the expected cost of a state-action pair $\bar{c}_i(s,a)=E_{P_{C,i}}[c|s,a]$ and a trajectory $\bar{c}_i(\tau)=\sum_{t=0}^\infty \bar{c}_i(s_t,a_t)$ when defining a constraint $E_{P_\pi}[\bar{c}_i(\tau)]\le\epsilon_i$.  Note that this type of constraint is generally 'soft' since it is possible that the cumulative cost of some trajectories exceeds the threshold $\epsilon_i$ as long as the expected cumulative cost remains bounded by $\epsilon_i$.  'Hard' constraints (i.e., when \emph{all} trajectories must individually have a cost bounded by $\epsilon_i$) can still be encoded for example by setting $\epsilon_i=0$ and making sure that costs are always non-negative.

The goal of constrained reinforcement learning (CRL) is to find an optimal policy that maximizes expected cumulative rewards subject to bounds on expected cumulative costs.
\begin{align}
& \pi^* = \arg \max_\pi E_{P_\pi} [\bar{r}(\tau)] \\
& \mbox{ such that } E_{P_\pi} [\bar{c}_i(\tau)]  \leq \epsilon_i \; \forall i \nonumber
\end{align}
Similarly, in the maximum entropy framework, an optimal policy $\pi^*_{maxent}$ is obtained by maximizing expected rewards with an entropy bonus subject to bounds on expected costs.
\begin{align}
& \pi^*_{maxent} = \arg \max_\pi E_{P_\pi} [\bar{r}(\tau)] + \beta H(P_\pi) \\
& \mbox{ such that } E_{P_\pi} [\bar{c}_i(\tau)]  \leq \epsilon_i \; \forall i \nonumber
\end{align}

\subsection{Inverse Constrained Reinforcement Learning}

In several applications, it is difficult to specify all the constraints required to find the most suitable optimal policy. However, it may be possible to obtain expert demonstrations that adhere to these underlying constraints. Hence, the goal in inverse constrained reinforcement learning (ICRL) \cite{malik2021inverse} is to make use of expert trajectories $D=\{\tau_j\}_{j=1}^n$ to recover the constraints. The ICRL framework is analogous to but differs from the relatively well-known inverse reinforcement learning (IRL) \cite{ng2000algorithms} framework. While IRL learns the reward function of an unconstrained MDP from expert demonstrations, ICRL learns the constraints in a constrained MDP under the assumption that the reward function $\bar{r}$ is available.

Several related prior works \cite{malik2021inverse, liu2023benchmarking} in ICRL recommend the use of the maximum entropy framework \cite{wu2012maximum}. In the case of hard constraints, the likelihood that an optimal policy $\pi^*_{maxent}$ will generate a trajectory $\tau$ is proportional to the exponential of the rewards times an indicator $\mathds{1}(\bar{c}_i(\tau)\le\epsilon_i \;\forall i)$. This is equal to 1 when the trajectory satisfies all constraints and 0 otherwise \cite{malik2021inverse}: 
\begin{align}
& P_{\pi^*_{maxent}}(\tau) = \frac{\exp(\beta \bar{r}(\tau)) \mathds{1}(\bar{c}_i(\tau)\le\epsilon_i \;\forall i)}{Z(\bar{c}_i)} \label{eq:trajectory-likelihood}\\
& \mbox{where } Z(\bar{c}_i) = \int_\tau \exp(\beta \bar{r}(\tau)) \mathds{1}(\bar{c}_i(\tau)\le\epsilon_i \;\forall i)d\tau \nonumber
\end{align}
If we replace the feasibility indicator $\mathds{1}(\bar{c}_i(\tau)\le\epsilon_i \;\forall i)$ by a differentiable neural network $\phi(\tau)$ with a sigmoid output and optimize its parameters to maximize the likelihood of the expert demonstrations $D$, we can learn an approximate feasibility indicator. Using this network, we will denote $Z(\phi)$ for $Z(\bar{c}_i)$.
\begin{equation}
\label{eq:max-likelihood}
\phi^*(\tau)= \arg \max_\phi \prod_{\tau\in D} \frac{\exp(\beta\bar{r}(\tau))\phi(\tau)}{Z(\phi)}
\end{equation}
In practice, $\phi(\tau)=\prod_t \phi(s_t,a_t)$ is often decomposed into a product of feasibility factors $\phi(s_t,a_t)$ for state-action pairs \cite{malik2021inverse}.  In the case of a single hard constraint, we can then set $\bar{c}(s,a)=1-\phi(s,a)$ and $\epsilon=0$ to obtain an equivalent constraint in the standard form.  This equivalence holds as long as $\phi(s,a)\in\{0,1\}$.  In practice, since $\phi(s,a)\in(0,1)$ (sigmoid outputs are never exactly 0 or 1), $\epsilon$ is typically set slightly above 0.  In the case of soft constraints, we can interpret $\phi(\tau)$ as an estimate of the probability that $\tau$ will be considered feasible \cite{liu2023benchmarking}.

Hence inverse constraint learning boils down to optimizing Eq.~\ref{eq:max-likelihood} which is done iteratively by alternating between forward control and constraint update \cite{malik2021inverse}.
\begin{enumerate}
\item Forward control: $\pi^* = \arg \max_\pi E_{P_\pi}[\bar{r}(\tau)]+\beta H(P_\pi)$ such that $E_{P_\pi}[\phi(\tau)]\le\epsilon$ \cite{ray2019benchmarking,gaurav2023learning}
\item Constraint update: $\phi \gets \phi +  \nabla_\phi  [\sum_{\tau\in D} \beta\bar{r}(\tau)) + \log \phi(\tau) - \log Z(\phi)]$  where the estimation of $Z(\phi)$ depends on $\pi^*$ \cite{malik2021inverse}
\end{enumerate}

Instead of estimating $\phi(\tau)$ by maximum likelihood based on (\ref{eq:max-likelihood}), one can also use Bayesian learning to estimate a posterior distribution $P(\phi|D)$ based on the set $D$ of expert trajectories by multiplying a suitable prior by the likelihood of each expert trajectory in Equation \ref{eq:trajectory-likelihood} \cite{Glazier2021Making, Papadimitriou2022Bayesian, liu2023benchmarking}.  However, since the resulting posterior does not have a closed form, it is projected in a tractable family of distributions~\cite{liu2023benchmarking} and the mean constraint is returned as a point estimate.

\section{Confidence Aware Inverse Constrained Reinforcement Learning}

As stated previously, we have two objectives in this paper. The first objective is inferring a constraint conditioned on a desirable confidence level.  Once such a constraint is inferred, the second objective consists of determining whether the number of expert trajectories used to infer the constraint is sufficient. In this section, we present the CA-ICRL algorithm, which is a principled ICRL algorithm that aims to satisfy both the stated objectives. For the first objective, the inputs consist of a set of expert trajectories that are assumed to be optimal in terms of maximizing rewards while satisfying some unknown underlying constraint, and a desired confidence level. The output consists of a constraint that is at least as constraining as the true underlying constraint with probability greater than or equal to the desired confidence level. For the second objective, the inputs consist of a constraint conditioned on a desired confidence level, and a desired reward level for an optimal policy that satisfies the constraint. The output is a Yes/No decision that indicates whether the number of expert trajectories are sufficient to find such a constraint that satisfies the desired confidence level and an associated policy that meets the desired reward level. When the number of expert trajectories are deemed insufficient, additional expert trajectories can be collected and added to the set of expert trajectories, based on which a looser constraint (that is still more constraining than the unknown underlying constraint with probability at least as great as a desired confidence level) can be inferred.  The process of collecting more expert trajectories and updating the inferred constraint continues until the number of expert trajectories is deemed sufficient.


\subsection{Objective 1: Inferring a constraint conditioned on a confidence level}

\begin{figure}
    \centering
    \includegraphics[width = 0.5\textwidth]{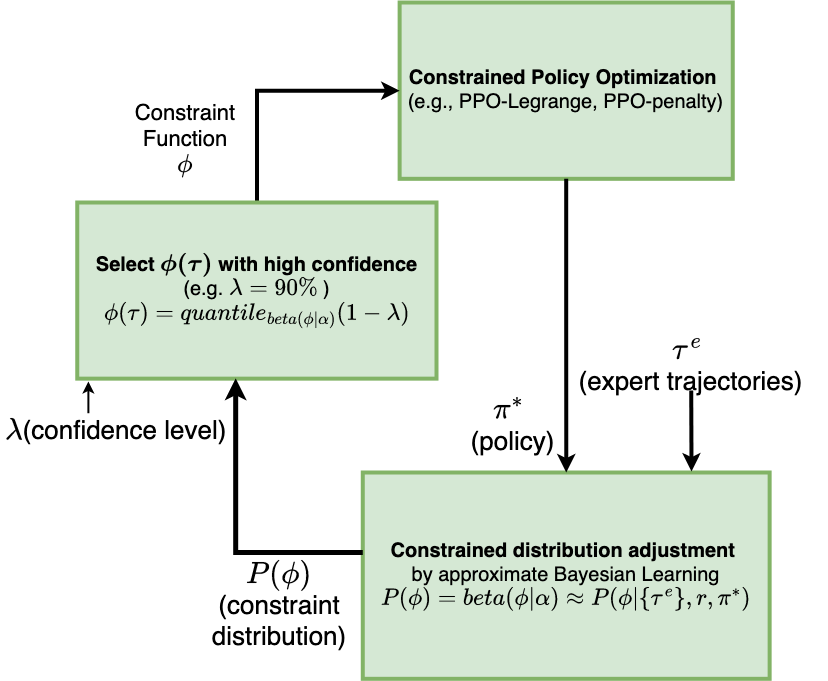}
    \caption{Solution for inferring confidence aware constraint from a set of expert trajectories with desired confidence $\lambda$}
    \label{fig:inferringconfidenceawaresolution}
\end{figure}

In Figure~\ref{fig:inferringconfidenceawaresolution} we provide an outline for the solution of our first objective that relates to inferring a confidence-aware constraint from a set of expert trajectories.  In prior methods \cite{Glazier2021Making, Papadimitriou2022Bayesian, liu2023benchmarking}, the distribution over constraints is used to compute an expectation over feasibility constraints $\E[\phi(\tau)] = \overline{\phi}(\tau)$. The resulting mean constraint $\overline{\phi}$ does not reflect any notion of confidence. Hence there is no guarantee that the mean constraint is at least as constraining as the unknown underlying constraint. Furthermore, the mean lacks sensitivity to the number of expert trajectories.  Normally, as the number of expert trajectories increases, we should be able to increase our confidence that a constraint is at least as constraining as the unknown underlying constraint. Alternatively, for a given confidence level, as we increase the number of trajectories, we should be able to infer a looser constraint that is still at least as constraining as the unknown constraint. Our solution achieves these objectives, but these were not possible in previous methods. 


To ease the understanding, let's start with an illustrative example.  Consider a lane change scenario in autonomous driving. Let $\tau = (s_0, a_0, s_1, a_1, \ldots, s_n, a_n)$  be a human driver trajectory where $s_t$ is the state at time step $t$, which includes the position and velocity of the ego car and surrounding cars.  Similarly, $a_t$ is the action at time step $t$, which includes the acceleration and steering of the ego car.
Let $\phi(\tau) \in [0,1]$ be the fraction of people who would judge the trajectory $\tau$ as safe.  More precisely, $\phi(\tau) = \prod_t \phi(s_t, a_t)$ could be decomposed into a product of feasibility factors $\phi(s_t, a_t)$ indicating the fraction of people who would judge the state-action pair at time $t$ as safe.  Note that $\phi(\tau)$ does not have to decompose into a product. This is simply an example for the case when the safety of a trajectory can be judged based on each state-action pair.  More generally, $\phi(\tau)$ can be any function that returns the probability that someone would judge $\tau$ as safe.

Let $P(\phi(\tau)) = beta(\phi(\tau) | \alpha)$  be the distribution over the fraction of people who would judge $\tau$ as safe. In this embodiment, this distribution is a Beta distribution with parameters $\alpha = [\alpha_1,\alpha_2]$. It represents the epistemic uncertainty of the learning algorithm, where epistemic uncertainty refers to the uncertainty that is due to a limited amount of data (e.g., limited number of expert trajectories).

Our choice of using the Beta distribution is not arbitrary. The beta distribution is being used to represent the distribution over the probability that the trajectory $\tau$ is safe. Here, our requirement is a continuous probability distribution that can represent a random variable with values falling inside a finite interval (in this case $[0,1]$). Note that the standard Beta distribution uses the interval $[0,1]$, which is ideal for modelling probabilities. This is why we go with the Beta distribution.  Particularly, our choice of the Beta distribution is motivated by its close relationship with the binomial distribution. The binomial distribution is used to model the number of successful outcomes in an experiment with binary outcomes (where each trial is a Bernoulli event). We are also considering a random variable, which is the outcome of an experiment having binary choices since the trajectory can be judged safe or unsafe (i.e., the two choices). The Beta distribution is widely used to model success in a binomial experiment, in terms of the binomial proportion, which is the fraction of the number of successes with respect to the total number of trials. Hence, we use the Beta distribution, which is an excellent choice to represent a distribution of probabilities.

Let $\lambda$ be a desired confidence level (e.g., 90\%) and let $\phi^*(\tau)$ be the highest fraction of people such that the true fraction of people $\phi(\tau)$ is at least as great as $\phi^*(\tau)$ with confidence $\lambda$ (i.e., $P(\phi(\tau) \geq \phi^*(\tau)) \geq \lambda$).  Hence, we wish to compute $\phi^*(\tau)$ since this is the feasibility constraint that corresponds to the confidence level $\lambda$. Here $\phi^*(\tau)$ corresponds to the $1-\lambda$ quantile of $beta(\phi(\tau) | \alpha)$.  By replacing $\phi(\tau)$ in Equation~\ref{eq:max-likelihood} with $quantile_{beta(\cdot| \alpha)} (1-\lambda)$, we obtain an optimization problem that allows us to learn the hyperparameters $\alpha$ of a beta distribution such that $\phi^*(\tau)$ is its $1-\lambda$ quantile.
\begin{equation}
max_{\alpha} \prod_{\tau\in D} \frac{\exp(\beta \bar{r}(\tau))quantile_{beta(\cdot|\alpha)}(1-\lambda)}{Z(\alpha)}
\end{equation}
In practice, we do not optimize $\alpha$ directly.  Instead we use a neural network that outputs $\alpha=[\alpha_1,\alpha_2]$ as depicted in Figure~\ref{fig:confidencearchitecture}.  Since this neural network takes as input the expert trajectories $D$ and a trajectory $\tau$ that we wish to evaluate for feasibility, we denote by $\alpha_w(D,\tau)$ the output of this neural network with weights $w$ (Figure~\ref{fig:confidencearchitecture} uses $W$ to denote the weights $w$).  The weights of this neural network are then optimized by using Equation~\ref{eq:error}. 
\begin{equation}\label{eq:error}
\max_{w} \prod_{\tau\in D} \frac{\exp(\beta \bar{r}(\tau))quantile_{beta(\cdot|\alpha_w(D,\tau))}(1-\lambda)}{Z(w)}
\end{equation}

The intuition behind the neural network in Figure~\ref{fig:confidencearchitecture} goes as follows.  Each encoder block shares the same set of weights $w$.  These encoder blocks can be bidirectional attention flows \cite{seo2016bidirectional}, transformers \cite{vaswani2017attention} or any other type of encoder that returns two numbers $\sigma^1_i$, $\sigma^2_i \in [0,1]$. $\sigma^1_i$ and $\sigma^2_i$ can be thought as the contribution of expert trajectory $\tau^e_i$ towards $\alpha_1$ and $\alpha_2$ in evaluating the feasibility of trajectory $\tau$.  In the beta distribution, $\alpha_1 - 1$ and $\alpha_2 -1$ can be interpreted as the number of data points of classes 1 and 2 respectively.  In our setting, the two classes are feasible and infeasible.  Hence, the outputs $\sigma^1_i$ and $\sigma^2_i$ of each encoder $i$ can be interpreted as fractional counts of feasibility and infeasibility when comparing a trajectory $\tau$ to an expert trajectory $\tau^e$ that is assumed to be feasible.

To summarize, by optimizing the weights $w$ to maximize $P(\phi(\tau))$, we are essentially trying to find weights that define a distribution over constraints $\phi(\tau)$ that ensures that the expert trajectories will be generated with a high probability.  The steps in confidence aware ICRL are given in Algorithm~\ref{alg:CA-ICRL}. 

\begin{algorithm}
\caption{Confidence Aware Inverse-Constraint-Learning}\label{alg:CA-ICRL}
\begin{algorithmic}
\REQUIRE Expert trajectories $D$, iterations $N$
\STATE  Initialize $\pi$ and $w$ randomly
\WHILE{$i = 1$ to $N$}
\STATE Forward control: 
\STATE $\quad \pi^* = \arg \max_\pi E_{P_\pi}[\bar{r}(\tau)]+\beta H(P_\pi)$ 
\STATE $\qquad$ such that $E_{P_C}[\phi^*(\tau)]\le\epsilon$
\STATE Update weights of constraint distribution:
\STATE $\quad w \gets w + \nabla_w  [\sum_{\tau\in D} \beta\bar{r}(\tau)] +$ 
\STATE $\qquad \log quantile_{beta(\cdot|\alpha_w(D,\tau))}(1-\lambda) - \log Z(w)]$
\STATE $\qquad$where estimation of $Z(w)$ depends on $\pi^*$
\STATE Compute $\phi^*$ based on confidence $\lambda$
\STATE $\quad \phi^*(\tau) \gets quantile_{beta(\cdot|\alpha_w(D,\tau))}(1-\lambda)$
\ENDWHILE
\end{algorithmic}
\end{algorithm}


\begin{figure}
    \centering
    \includegraphics[width = 0.5\textwidth]{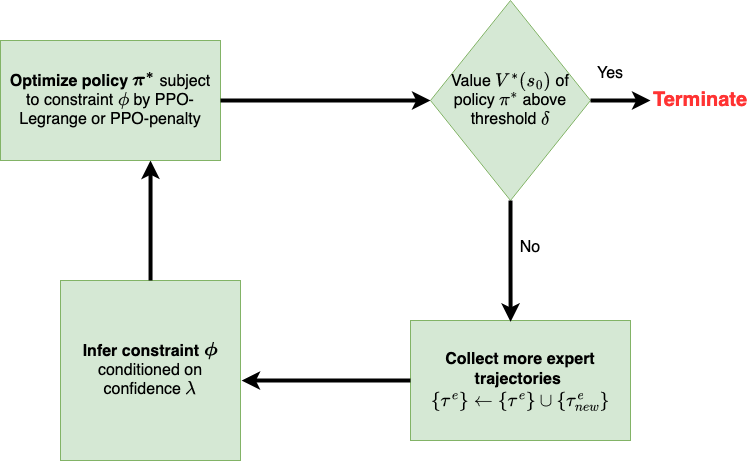}
    \caption{Solution for determining sufficiency of expert trajectory based on the given confidence $\lambda$}
    \label{fig:determiningsufficiencysolution}
\end{figure}

\subsection{Objective 2: Determining sufficiency of expert trajectories}\label{sec:sufficiency}

Figure~\ref{fig:determiningsufficiencysolution} outlines the solution for the second problem about determining whether the number of expert trajectories is sufficient or not.   Note that the bottom left box in Figure~\ref{fig:determiningsufficiencysolution} corresponds to all of Figure~\ref{fig:inferringconfidenceawaresolution}.  Hence, we can view Figure~\ref{fig:inferringconfidenceawaresolution} as the step of inferring a confidence-aware constraint in Figure~\ref{fig:determiningsufficiencysolution}.  This solution allows practitioners to determine whether they have enough expert trajectories.  Intuitively, when the number of expert trajectories is small, then the inferred constraint will be very constraining, yielding a policy of low value.  As the number of expert trajectories increases, the inferred constraint for a given confidence level can be relaxed, yielding a policy of increased value.  Hence, for a given confidence level, the process in Figure~\ref{fig:inferringconfidenceawaresolution} will increase the number of expert trajectories until we get a policy that exceeds a desired value threshold.

\begin{figure}
    \centering
    \includegraphics[width=8cm]{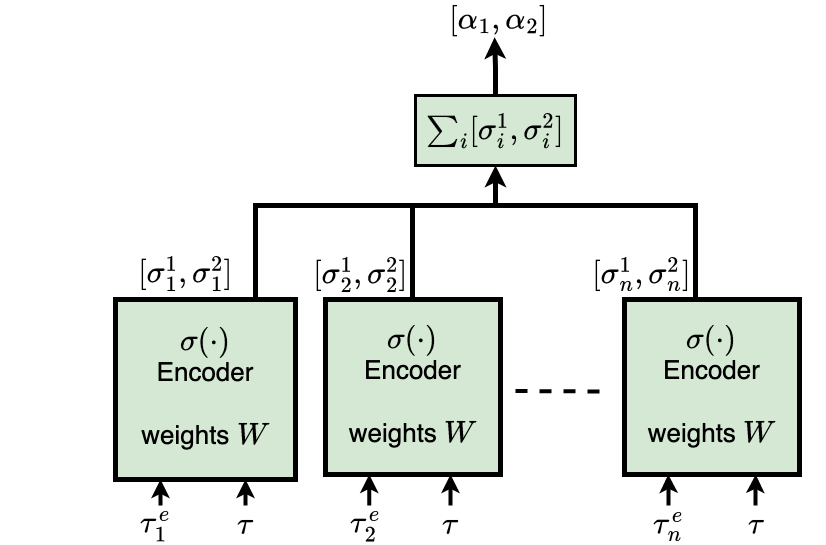}
    \caption{Confidence Architecture}
    \label{fig:confidencearchitecture}
\end{figure}

\section{Experiments and Results}\label{sec:experiments}

We follow the same procedure as prior works~\cite{malik2021inverse, liu2023benchmarking} for evaluating the ICRL algorithms. The objective is to learn a policy that obtains as much rewards as possible by adhering to the constraints. Hence, we plot the constraint violation rate as well as the rewards obtained by different ICRL methods to demonstrate performances and perform comparisons. The \emph{constraint violation rate} provides the probability with which a constraint is violated in a trajectory by a policy (lower is better). The \emph{rewards} or cumulative rewards provides the total rewards collected by an agent in trajectories without constraint violation (higher is better).
We use four baselines for comparison. Binary Classifier Constraint Learning (BC2L)~\cite{liu2023benchmarking}, Generative Adversarial Constraint Learning (GACL)~\cite{Ho2016Generative}, Inverse Constrained Reinforcement Learning (ICRL)~\cite{malik2021inverse}, and Variational Inverse Constrained Reinforcement Learning (VICRL) \cite{liu2023benchmarking}. While the baselines are predominantly algorithms that learn constraints from data, the GACL method is an IRL algorithm (specifically GAIL from \citet{Ho2016Generative}) that has been adapted to the ICRL setting by directly modifying the reward function to provide large punishments for violating the constraints (see \citet{liu2023benchmarking} for more details on this algorithm). We use two types of environments for comparing empirical performances. The first is a set of virtual environments from the well-known MuJoCo \cite{Todorov2012International} simulator. The second is a realistic environment based on a highway driving task previously used by \citet{liu2023benchmarking}. We consider a total of seven domains for the experiments (five within Mujoco and two on the highway driving task). 

All experiments are repeated 50 times and we report the average and standard deviation of performances. Further, we conduct an unpaired 2-sided t-test and report $p$-values for statistical significance. As is common in literature, we will consider $p<0.05$ as statistically significant differences. 
All experiments are conducted in two phases. The first phase is training, where CA-ICRL and all baselines train the constraint adjustment and policy update networks over a set of training episodes. The second phase is testing or execution, where there is no further training and no exploratory moves. Each method simply executes its learned policies from the training phase. We report results across both phases.  All the code for the experiments have been open-sourced~\cite{sourcecode}.

\subsection{MuJoCo (Stochastic) Virtual Environments}

The default MuJoCo environments commonly used for RL have been modified for constraint inference by prior works~\cite{liu2023benchmarking, Baert2023Maximum}. We use similar environments for our empirical studies as well. 
We consider 5 different robotic environments from MuJoCo (stochastic having Guassian noise in the transitions with $\sigma = 0.2$, see Appendix~\ref{appendix:implementation}). These are the well-known Half Cheetah, Ant, Pendulumn, Walker and Swimmer environments~\cite{Duan2016benchmarking}. They have been modified by adding constraints and making corresponding adjustments to the reward function (details in Appendix~\ref{appendix:implementation}). 
Following prior work~\cite{malik2021inverse, liu2023benchmarking}, we generate an expert dataset by training a PPO-Lagrange method~\cite{ray2019benchmarking} with the ground truth constraints, and running the trained PPO-Lagrange expert agent in testing environments where trajectories that do not violate the provided constraints are added to a data buffer $\mathcal{D}_e$.

\begin{figure*}
\centering
\begin{subfigure}{.18\textwidth}
  \centering
\includegraphics[width=3cm, height = 3cm]{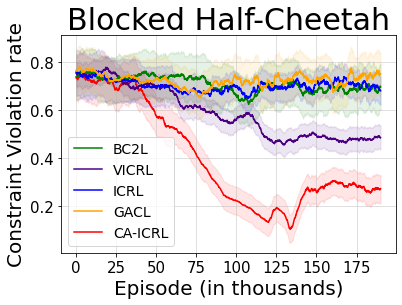}
\label{fig:Blockedhalfcheetahconstraintviolation}
\end{subfigure}%
\begin{subfigure}{.18\textwidth}
  \centering
  \includegraphics[width=3cm, height = 3cm]{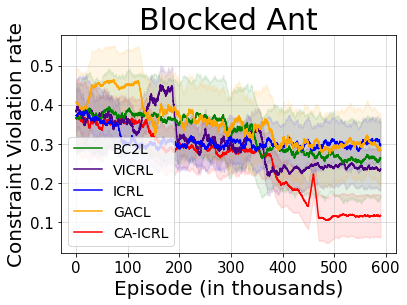}
  \label{fig:Blockedantconstraintviolation}
\end{subfigure}
\begin{subfigure}{.18\textwidth}
  \centering
  \includegraphics[width=3cm, height = 3cm]{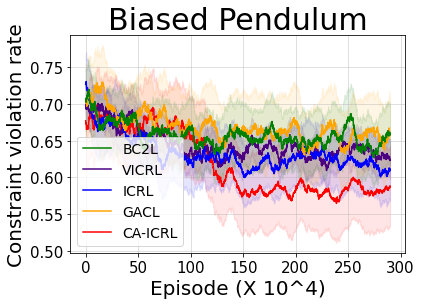}
  \label{fig:Biasedpendulumconstraintviolation}
\end{subfigure}
\begin{subfigure}{.18\textwidth}
  \centering
  \includegraphics[width=3cm, height = 3cm]{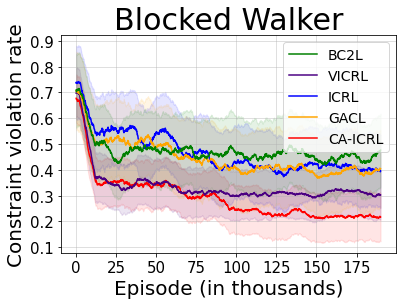}
  \label{fig:blockedwalkerconstraintviolation}
\end{subfigure}
\begin{subfigure}{.18\textwidth}
  \centering
  \includegraphics[width=3cm, height = 3cm]{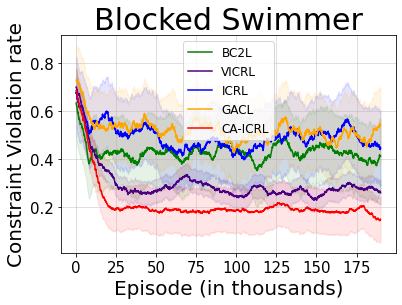}
  \label{fig:blockedswimmerconstraintviolation}
\end{subfigure}

\begin{subfigure}{.18\textwidth}
  \centering
\includegraphics[width=3cm, height = 3cm]{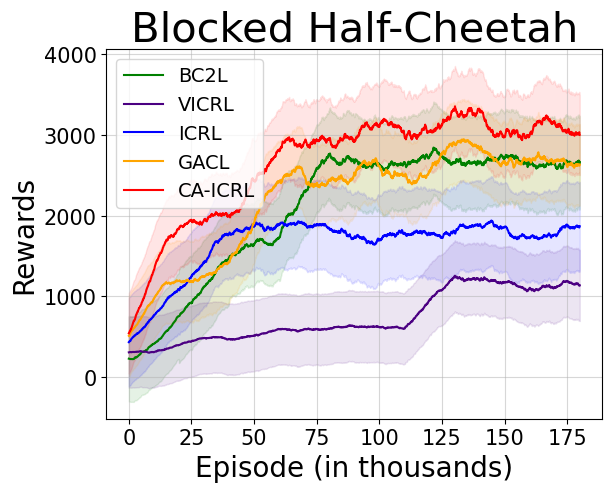}
\label{fig:Blockedhalfcheetahfeasiblerewards}
\end{subfigure}%
\begin{subfigure}{.18\textwidth}
  \centering
  \includegraphics[width=3cm, height = 3cm]{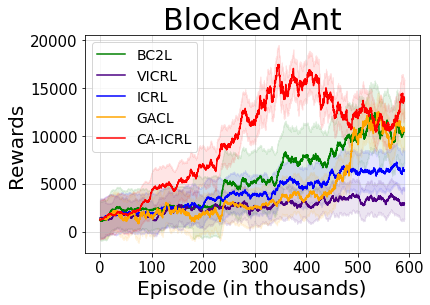}
  \label{fig:Blockedantfeasiblerewards}
\end{subfigure}
\begin{subfigure}{.18\textwidth}
  \centering
  \includegraphics[width=3cm, height = 3cm]{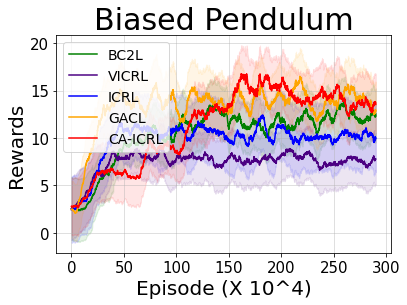}
  \label{fig:Biasedpendulumfeasiblerewards}
\end{subfigure}
\begin{subfigure}{.18\textwidth}
  \centering
  \includegraphics[width=3cm, height = 3cm]{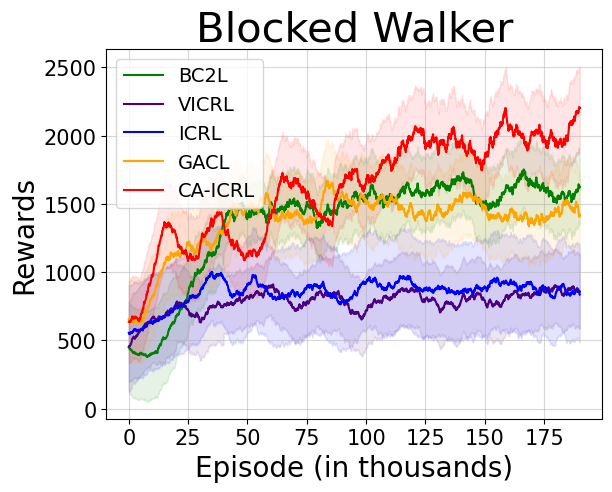}
  \label{fig:blockedwalkerfeasiblerewards}
\end{subfigure}
\begin{subfigure}{.18\textwidth}
  \centering
  \includegraphics[width=3cm, height = 3cm]{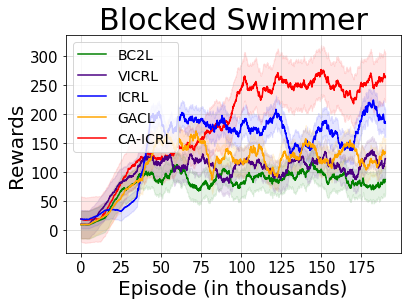}
  \label{fig:blockedswimmerfeasiblerewards}
\end{subfigure}

\caption{Training experiments in the MuJoCo environments}
\label{fig:constraintviolationratetraining}
\end{figure*}
\begin{figure*}
\centering
\begin{subfigure}{.18\textwidth}
  \centering
\includegraphics[width=3cm, height = 3cm]{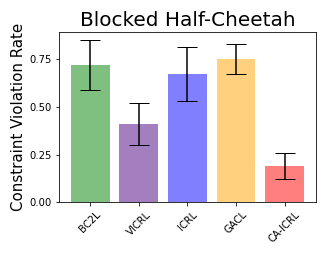}
\label{fig:Blockedhalfcheetahconstraintviolationexecution}
\end{subfigure}%
\begin{subfigure}{.18\textwidth}
  \centering
  \includegraphics[width=3cm, height = 3cm]{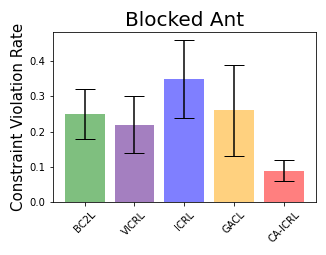}
  \label{fig:Blockedantconstraintviolationexecution}
\end{subfigure}
\begin{subfigure}{.18\textwidth}
  \centering
  \includegraphics[width=3cm, height = 3cm]{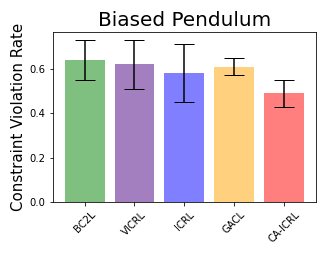}
  \label{fig:Biasedpendulumconstraintviolationexecution}
\end{subfigure}
\begin{subfigure}{.18\textwidth}
  \centering
  \includegraphics[width=3cm, height = 3cm]{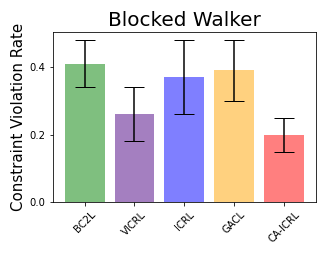}
  \label{fig:blockedwalkerconstraintviolationexecution}
\end{subfigure}
\begin{subfigure}{.18\textwidth}
  \centering
  \includegraphics[width=3cm, height = 3cm]{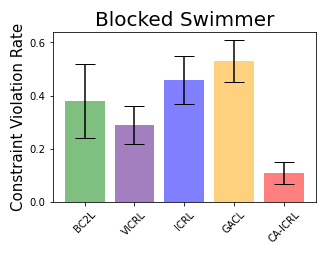}
  \label{fig:blockedswimmerconstraintviolationexecution}
\end{subfigure}

\begin{subfigure}{.18\textwidth}
  \centering
\includegraphics[width=3cm, height = 3cm]{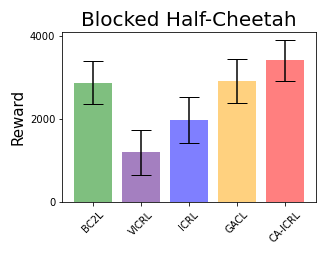}
\label{fig:Blockedhalfcheetahfeasiblerewardsexecution}
\end{subfigure}%
\begin{subfigure}{.18\textwidth}
  \centering
  \includegraphics[width=3cm, height = 3cm]{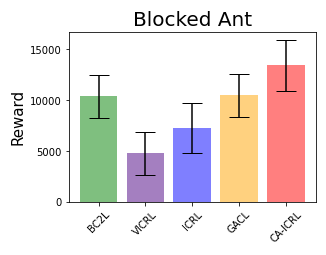}
  \label{fig:Blockedantfeasiblerewardsexecution}
\end{subfigure}
\begin{subfigure}{.18\textwidth}
  \centering
  \includegraphics[width=3cm, height = 3cm]{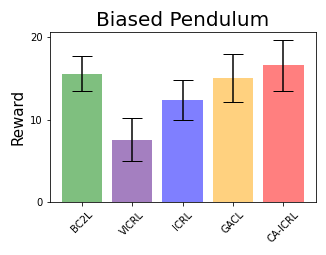}
  \label{fig:Biasedpendulumfeasiblerewardsexecution}
\end{subfigure}
\begin{subfigure}{.18\textwidth}
  \centering
  \includegraphics[width=3cm, height = 3cm]{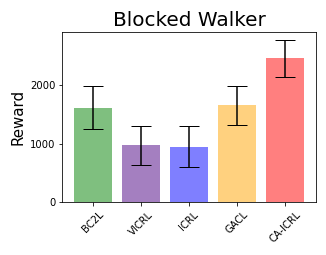}
  \label{fig:blockedwalkerfeasiblerewardsexecution}
\end{subfigure}
\begin{subfigure}{.18\textwidth}
  \centering
  \includegraphics[width=3cm, height = 3cm]{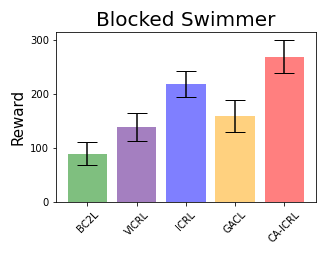}
  \label{fig:blockedswimmerfeasiblerewardsexecution}
\end{subfigure}

\caption{Execution experiments in the MuJoCo environments}
\label{fig:constraintviolationrateexecution}
\end{figure*}

Figure~\ref{fig:constraintviolationratetraining} shows the constrained violation rate and rewards obtained by the different methods in all the MuJoCo environments during the training phase. For the CA-ICRL method we select a confidence value of 70\%. This means that CA-ICRL is learning constraints that are at-least as constraining as the ground truth constraints 70\% of the times. From the constraint violation rate plots in Figure~\ref{fig:constraintviolationratetraining}, we can see that the constraint violation rate does indeed fall below 30\% for all environments except the Biased Pendulum environment. The Biased Pendulum is a hard environment where all ICRL methods struggle to achieve good performances (in both the violation rate and rewards obtained), as also observed by prior works~\cite{liu2023benchmarking, Baert2023Maximum}. However, the CA-ICRL method still does better than the other baselines by achieving a lesser constraint violation rate as shown in Figure~\ref{fig:constraintviolationratetraining}. 
Further, from Figure~\ref{fig:constraintviolationratetraining} we can see that the CA-ICRL method consistently obtains more rewards compared to other methods during training. 
CA-ICRL learns a constraint function that is at-least as constraining as the ground truth constraints with a desired level of confidence, while the other methods learn a constraint function without a notion of confidence. Having a high confidence threshold (such as 70\%) makes CA-ICRL more conservative right from the beginning of training. Further on in training, this conservative behaviour needs fewer adjustments to learn a reasonable constraint function and associated policy. Comparatively, other methods start with a policy and a constraint function that are initially aggressive (since the policy is trying to maximize rewards and the learned constraints are poor, the methods become aggressive by default), violating constraints many times before learning a reasonable constraint function and policy. This means that these methods have to progressively unlearn their aggressive behaviour, before they can learn good policies.  Another advantage of CA-ICRL is that, based on need, the confidence threshold can be reduced to make the method more aggressive if required (which is not possible in other methods). The relative advantage of CA-ICRL in the different MuJoCo environments can also be observed in the testing performances as seen in Figure~\ref{fig:constraintviolationrateexecution}. The observations are statistically significant (refer Appendix~\ref{appendix:additionalresults}). 

In our experimental domains in Figure~\ref{fig:constraintviolationratetraining}, we found that confidence values in the range of 70\% -- 80\% were most ideal for the performance of CA-ICRL to balance the twin goals of having a low constraint violation rate and the rewards obtained. Nonetheless, CA-ICRL can be used with any value of confidence that is usually influenced by practical considerations.

\begin{figure*}
\centering
\begin{subfigure}{.33\textwidth}
  \centering
\includegraphics[width=3cm, height = 3cm]{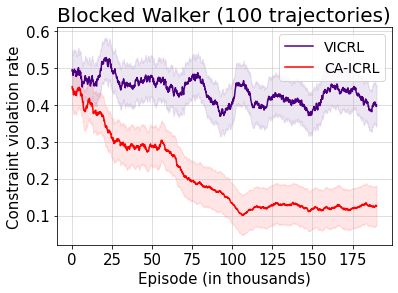}

\end{subfigure}%
\begin{subfigure}{.33\textwidth}
  \centering
  \includegraphics[width=3cm, height = 3cm]{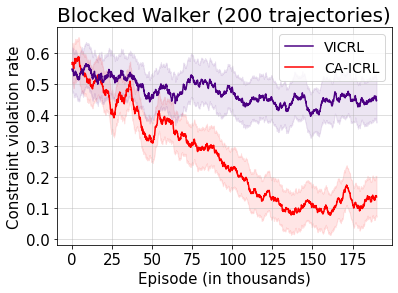}
 
\end{subfigure}
\begin{subfigure}{.33\textwidth}
  \centering
  \includegraphics[width=3cm, height = 3cm]{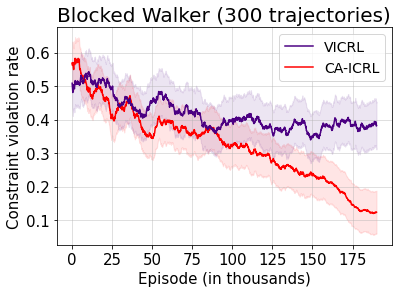} 
  
\end{subfigure}

\begin{subfigure}{.33\textwidth}
  \centering
\includegraphics[width=3cm, height = 3cm]{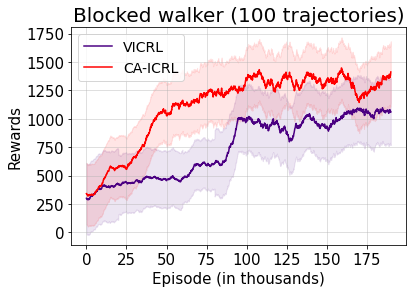}

\end{subfigure}%
\begin{subfigure}{.33\textwidth}
  \centering
  \includegraphics[width=3cm, height = 3cm]{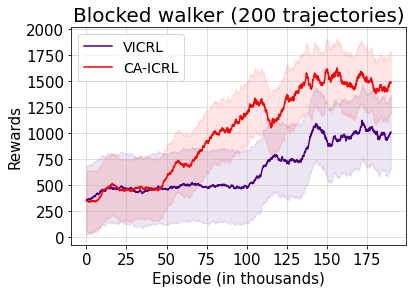}
 
\end{subfigure}
\begin{subfigure}{.33\textwidth}
  \centering
  \includegraphics[width=3cm, height = 3cm]{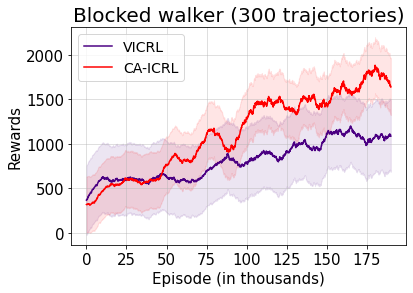} 
  
\end{subfigure}

\caption{Comparison of VICRL and CA-ICRL in the Blocked Walker Environment for different numbers of expert trajectories}
\label{fig:blockedwalkerdifferenttrajectoriesconstraintviolationrate}
\end{figure*}



\begin{figure*}
\centering
\begin{subfigure}{.24\textwidth}
  \centering
\includegraphics[width=3cm, height = 3cm]{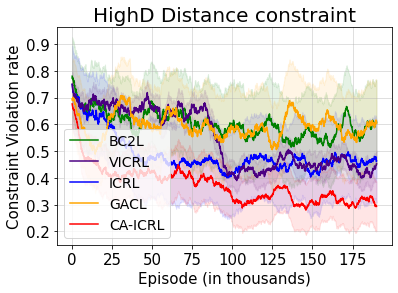}
\end{subfigure}%
\begin{subfigure}{.24\textwidth}
  \centering
  \includegraphics[width=3cm, height = 3cm]{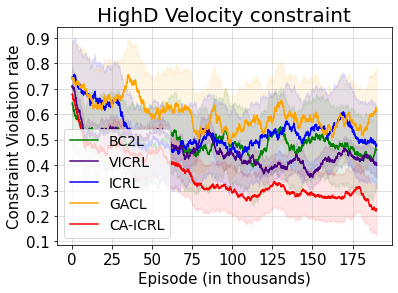}
\end{subfigure}
\begin{subfigure}{.24\textwidth}
  \centering
  \includegraphics[width=3cm, height = 3cm]{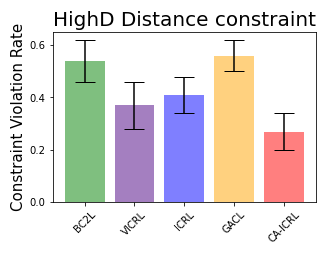}
\end{subfigure}
\begin{subfigure}{.24\textwidth}
  \centering
  \includegraphics[width=3cm, height = 3cm]{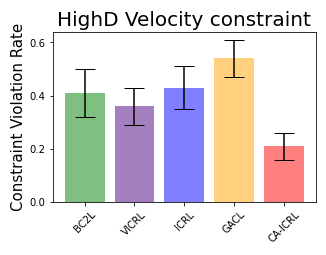}
\end{subfigure}

\begin{subfigure}{.24\textwidth}
  \centering
\includegraphics[width=3cm, height = 3cm]{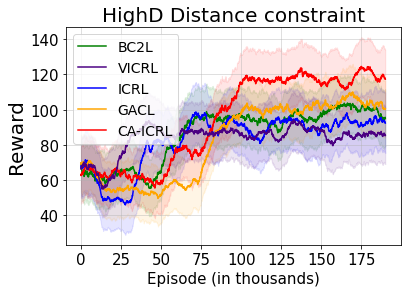}
\end{subfigure}%
\begin{subfigure}{.24\textwidth}
  \centering
  \includegraphics[width=3cm, height = 3cm]{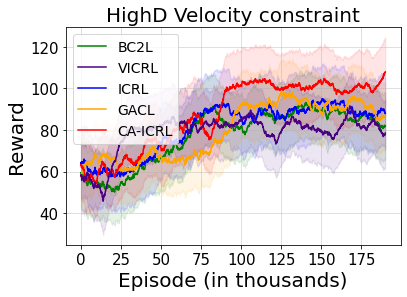}
\end{subfigure}
\begin{subfigure}{.24\textwidth}
  \centering
  \includegraphics[width=3cm, height = 3cm]{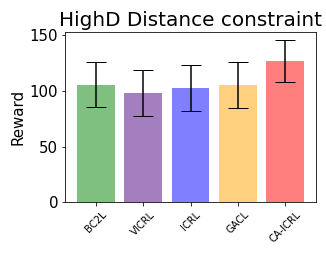}
\end{subfigure}
\begin{subfigure}{.24\textwidth}
  \centering
  \includegraphics[width=3cm, height = 3cm]{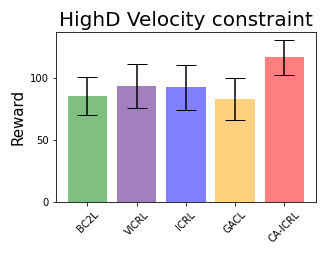}
\end{subfigure}

\caption{Training and execution in the HighD environments with distance and velocity constraints}
\label{fig:highDconstraintviolation}
\end{figure*}

\begin{figure}
  \begin{subfigure}[b]{0.44\columnwidth}
    \includegraphics[width=\linewidth, height=4cm]{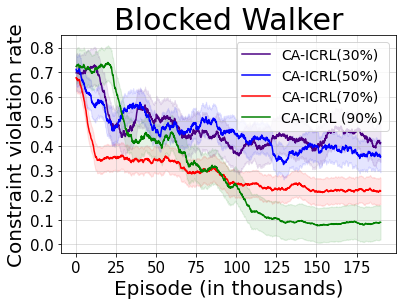}
  \end{subfigure}
  \hfill 
  \begin{subfigure}[b]{0.44\columnwidth}
    \includegraphics[width=\linewidth, height=4cm]{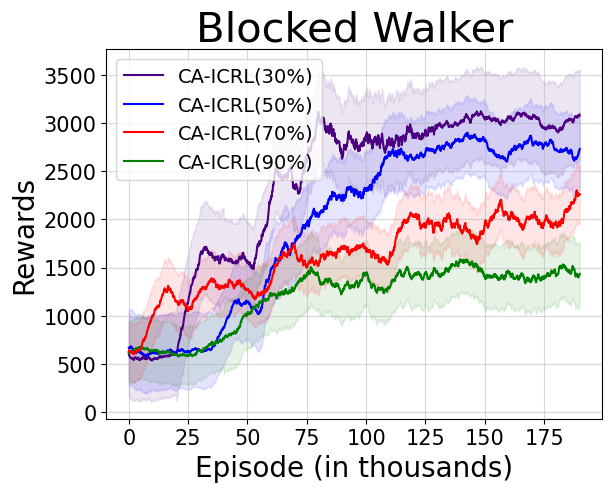}
  \end{subfigure}
  \caption{CA-ICRL performance with varying confidence values}
\label{fig:blockedwalkervaryingconfidencevalues}
\end{figure}


To demonstrate our second objective (i.e., to show that CA-ICRL allows practitioners to know if the number of expert trajectories are sufficient for the given task when provided a desired level of confidence and rewards), we conduct another set of experiments where we vary the number of expert trajectories and plot the performances of CA-ICRL. In this scenario, since practitioners have a desired confidence and performance level they need to meet, only ICRL methods that learn a distribution over constraints can be considered. The ICRL methods that learn point estimates (single constraint estimate) cannot provide confidence measures and hence are not considered as part of this experiment. From our set of algorithms, only VICRL and CA-ICRL learn distribution over constraints. Hence, only these two algorithms are considered as part of this experiment.

For these experiments, we assume a confidence of $80\%$ for CA-ICRL. We consider three scenarios, where there are 100, 200, and 300 expert trajectories available respectively. The performances in Blocked Walker are in Figure~\ref{fig:blockedwalkerdifferenttrajectoriesconstraintviolationrate} (other environments can be found in Appendix~\ref{appendix:additionalresults}). It can be observed that CA-ICRL obtains a much better constraint violation rate as compared to VICRL. Further, in all the three cases, the constraint violation rate of CA-ICRL eventually drops to less than $20\%$, consistent with the confidence requirement of $80\%$. 
Also, we observe that the rewards obtained by CA-ICRL is higher than that obtained by VICRL across all the three different scenarios. Additionally, CA-ICRL obtains higher rewards as the number of expert trajectories increases. This shows that CA-ICRL learns less conservative constraints when more expert trajectories are present, and practitioners can use CA-ICRL to determine if the number of expert trajectories are sufficient to achieve the desired performance levels (i.e., they will need to obtain more trajectories if CA-ICRL does not learn policies that reach desired performance levels with the current trajectories). 

\subsection{Realistic Environments}

This environment corresponds to the highway driving task used by prior works \cite{liu2023benchmarking, Baert2023Maximum}. Here we have a set of human driver trajectories collected as part of the HighD dataset~\cite{Krajewski2018high}. The HighD dataset collects samples reflecting the stochastic dynamics involved in real-world driving (due to different preferences of human drivers and road conditions). This dataset preparation and the game conditions are the same as that in \citet{liu2023benchmarking}, where the velocity and distance of vehicles are constrained to ensure a safe driving experience. Two environments are simulated using this dataset, where the first environment uses a constraint of the vehicle distance and the second environment places a constraint on the vehicle velocity (more details in Appendix~\ref{appendix:implementation}).  
The performances of the different algorithms are in Figure~\ref{fig:highDconstraintviolation} (CA-ICRL continues to use the confidence of 70\%). Similar to our observations in MuJoCo, we find that CA-ICRL provides better performances as compared to other methods in terms of both constraint violation rate and rewards. 



\subsection{Varying Confidence in CA-ICRL}

We run an experiment to study the performance of CA-ICRL under different confidence values. As noted previously, one important advantage of CA-ICRL as compared to prior ICRL methods is that it can accept a confidence value and learn the least constraining constraint that is at-least as constraining as the ground truth constraints with the desired confidence. We choose the Blocked Walker MuJoCo environment and four confidence values (30\%, 50\%, 70\%, and 90\%) for CA-ICRL. For each confidence value we use a expert dataset with 150 trajectories. We plot the training performances (both constraint violation rate and the rewards obtained) in Figure~\ref{fig:blockedwalkervaryingconfidencevalues}. From results in Figure~\ref{fig:blockedwalkervaryingconfidencevalues}(a), we note that as the confidence values increases, the constraint violation rate of CA-ICRL reduces. Hence, if practitioners prefer to be conservative and only use constraints that have high confidences, then the constraint violation rate of CA-ICRL drops as expected. However, from Figure~\ref{fig:blockedwalkervaryingconfidencevalues}(b), we can see that the performance in terms of reward accumulated also reduces with increase in confidence. This is also expected, since requiring more confidence in the constraints leads to learning comparatively more conservative constraints that reduces the rewards obtained. Considering this trade-off between rewards and the constraint violation rate, practitioners can use CA-ICRL with the choice of an appropriate confidence based on their requirements. Further, Figure~\ref{fig:blockedwalkervaryingconfidencevalues}(a) demonstrates that CA-ICRL ensures that the constraints are at-least as conservative as the ground truth constraints with the desired value of confidence for all the four values of confidence. This demonstrates the advantage of CA-ICRL as compared to prior methods that do not maintain confidence. 

In our work, we discussed two ways of meeting the performance requirements (represented by $\delta$) for the CA-ICRL algorithm. Either, the number of expert trajectories can be increased by fixing the confidence threshold (i.e., $\lambda$) as shown in the experiments of Figure~\ref{fig:blockedwalkerdifferenttrajectoriesconstraintviolationrate}, or the confidence requirements can be reduced by keeping the expert trajectories fixed as shown in Figure~\ref{fig:blockedwalkervaryingconfidencevalues}. Based on the preferences of the practitioners and their performance requirements, either of the two methods can be used. 

Appendix~\ref{appendix:wallclocktimes} provides the wall clock time for our experiments, and Appendix~\ref{appendix:computationalcomplexity} contains details regarding the computational complexity for CA-ICRL. 

\subsection{Minimum Number of Expert Trajectories}

\begin{figure}
    \centering
    \includegraphics[width=\linewidth, height=5.2cm]{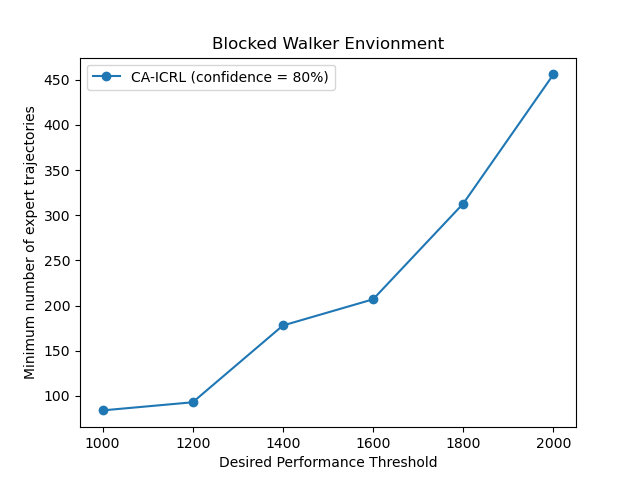}
    \caption{Experiment showing the minimum number of expert trajectories required for a desired value threshold in CA-ICRL. The experiment uses the Blocked Walker environment. CA-ICRL uses a confidence of 80\%.}
    \label{fig:blockedwalkerablation}
\end{figure}

In Figure~\ref{fig:blockedwalkerdifferenttrajectoriesconstraintviolationrate} we showed that the performance of CA-ICRL improves with an increase in the number of expert trajectories for a fixed confidence requirement. Related to this observation, in this subsection, we run an ablation study that demonstrates an increase in the minimum numbers of expert trajectories required for CA-ICRL to satisfy increases in performance requirements (represented by $\delta$ in Section~\ref{sec:sufficiency}). We consider the Blocked Walker environment, where we vary the performance threshold ($\delta$) from 1000 to 2000 (increments of 200) and use CA-ICRL (with a confidence requirement of 80\%) to find the minimum number of expert trajectories from the expert policy to achieve this performance. The ablation results can be found in Figure~\ref{fig:blockedwalkerablation}. The results show that more trajectories are required for a higher desired performance, consistent with our observations in other experiments in the main paper.

\section{Conclusion}

In this paper, we introduced the notion of confidence in ICRL. We provided a method (CA-ICRL) that can take in a desired level of confidence, and learn a constraint that is at-least as constraining as the ground truth constraint with the desired level of confidence. Additionally, this method allows practitioners to know if the number of expert trajectories available is sufficient to learn constraints with the desired confidence and performance levels. Empirically, we demonstrated the performance of CA-ICRL in a variety of simulated and realistic environments and showcased its superiority as compared to a set of recent baselines. A set of limitations of our work with associated discussion of future works can be found in Appendix~\ref{appendix:limitations}. Particularly, as future work, we are interested in extending our study to settings where the constraint as well as the reward function are not provided apriori and need to be learned from data.

\section*{Acknowledgements}

Resources used in preparing this research at the University of Waterloo were provided by Huawei Canada, NSERC, the province of Ontario and the government of Canada through CIFAR and companies sponsoring the Vector Institute. 

\section*{Impact Statement}

This work provides an algorithm in the space of Inverse Constrained Reinforcement Learning (ICRL), that maintains a measure of confidence in the estimated constraints from a set of expert demonstrations. We expect our method to have a very positive impact in the area of AI safety. To deploy algorithms that are safe, practitioners can use CA-ICRL with a high confidence requirement. This is extremely important in safety critical applications like autonomous driving. For example in the case of autonomous driving, if we pick a confidence of $90\%$, CA-ICRL can be used to find the minimum gap $d$ (bumper-to-bumper distance) such that the true minimum gap that is safe in practice is at-least $d$, $90\%$ of the time.  

\bibliography{ICML/submission}
\bibliographystyle{icml2024}

\newpage
\appendix
\onecolumn

\section{Limitations and Future Work}\label{appendix:limitations}

We can think of a few limitations of our work. These could serve as excellent avenues for future work. 

\begin{enumerate}
    \item Our algorithm like several other ICRL methods~\cite{liu2023benchmarking, Baert2023Maximum, gaurav2023learning} assumes that the reward function is available and only the constraint function needs to be learned. Learning the constraint function as well as the reward function from expert demonstrations is an important avenue for future work. 
    \item ICRL methods (including ours) commonly learn the imitation policy using online training with a simulator~\cite{liu2023benchmarking}. However, in many real-world settings like autonomous driving, it is hard to build a simulator that simulates the conditions perfectly. Rather, it is easier to collect offline data and we need ICRL methods that can learn the imitation policy through offline training. 
    \item It is commonly assumed that the expert demonstration used for training ICRL methods are coming from oracles that are infallible. However, in real-world settings it is unlikely to have such experts/oracles who are infallible. Extensions of ICRL methods to sub-optimal experts is another important avenue for future work. 
    \item Even though our method provides a way to use confidence associated with a learned policy in ICRL, there is still no way to guarantee that this policy will not violate ground-truth constraints. Learning such policies that guarantee zero ground truth constraint violation rate is yet another important avenue for future work. 
    \item Our method is slower than previous methods since it learns and uses a similarity measure between every policy trajectory from the current policy and all expert trajectories to determine the revised constraint function. This method does not scale well to problems with an extremely large set of expert trajectories. 
\end{enumerate}

\section{Related Work}\label{appendix:relatedwork}

\begin{table*}[h!]
\centering
\begin{tabular}{||p{5.5cm} | p{2cm} | p{1.8cm} | p{2.5cm} | p{3cm} ||} 
 \hline \hline
 Algorithms & State-Action Space & Soft/Hard constraints & Distribution over constraints & Confidence-aware constraints  \\ 
 \hline 
 ML-ICL \cite{dexter2020maximum} & Discrete & Hard & No & No  \\ 
 \hline
 ML-SICL \cite{McPherson2021Maximum}  & Discrete & Hard & No & No \\ 
 \hline
 ICLR \cite{malik2021inverse} & Continuous & Hard & No & No  \\ 
 \hline
 GAIL \cite{Ho2016Generative} & Continuous & Hard & No & No  \\ 
 \hline
 P-ICL \cite{Chou2019Learning} & Continuous & Hard & No & No \\ 
 \hline
 Soft-ICL \cite{gaurav2023learning} & Continuous & Soft & No & No  \\ 
 \hline
 UA-ICL \cite{Chou2020Uncertainty} & Continuous & Hard & Yes & No  \\ 
 \hline
 MESC-IRL \cite{Glazier2021Making} & Discrete & Soft & Yes & No \\ 
\hline
 BICRL \cite{Papadimitriou2022Bayesian} & Discrete & Hard & Yes & No   \\ 
 \hline
 VICRL \cite{liu2023benchmarking} & Continuous & Soft & Yes & No  \\ 
 \hline
 CA-ICRL (ours) & Continuous & Soft & Yes & Yes \\ 
 \hline
\end{tabular}
\caption{State-of-the-art ICRL algorithms compared. Only CA-ICRL learns a confidence-aware constraint.}
\label{tab:ICRLalgorithms}
\end{table*}

Previous approaches for inverse constrained learning can be broadly classified based on the type of state-action space (discrete/continuous) they are applicable to, type of constraints they learn (soft/hard), whether they learn a distribution over constraints, and whether the constraints learned are associated with a confidence level. The difference between soft and hard constraints is that the hard constraints need to be strictly satisfied in every trajectory, while the soft constraints only need to hold in expectation \cite{gaurav2023learning}. Table~\ref{tab:ICRLalgorithms} captures these characteristics of prior methods in the literature. The majority of the works that restrict themselves to discrete states and actions focus on learning constraints that evidently distinguish between feasible and infeasible state-action pairs \cite{dexter2020maximum, McPherson2021Maximum, Park2019Inferring}. Analogously,  other works that are applicable to continuous domains learn a decision boundary to distinguish between feasible and infeasible regions of state-action pairs \cite{malik2021inverse, Chou2019Learning, Lin2017Learning}. While studying continuous domains, all works are motivated by specific applications in the field of robotics. \citet{Armesto2017Efficient} learns constraints directly from observations associated with a surface wiping task pertaining to robotic arm movements.  \citet{Arpino2017learning} learns geometric constraints and associated policies for reaching and grasping movements with a robotic arm. \citet{Menner2021constrained} learns the objective function and constraints regarding predictive models of human motor control. These previous works have all assumed the availability of the transition model, so that planning can be undertaken using the learned constraints to find the optimal policy. In contrast, \citet{malik2021inverse, gaurav2023learning, liu2023benchmarking} have provided algorithms that can learn constraints and associated policies in a fully model-free setting with possibly high-dimensional state-action space continuous environments. Out of these methods, \citet{gaurav2023learning, liu2023benchmarking} aims to learn soft constraints, while \citet{malik2021inverse} learn hard constraints. All of these methods use neural networks to learn the constraints and the associated policies. Previously, ICRL methods were mostly restricted to learning a single candidate constraint \cite{dexter2020maximum, McPherson2021Maximum, malik2021inverse}, while more recent works learn a distribution over constraints \cite{Chou2020Uncertainty, Papadimitriou2022Bayesian, liu2023benchmarking}. In these methods, \citet{Chou2020Uncertainty} also follows the planning paradigm, where the learned constraints are used with the available transition functions to find the optimal policy. In contrast, \citet{Papadimitriou2022Bayesian} uses reinforcement learning to learn the optimal policy without assuming access to the environment's transition model, however this method only applies to discrete state-action space environments. Along similar lines, \citet{Glazier2021Making} learns a distribution over constraints with a scaling factor applied to rewards following a logistic distribution, with the method restricted to discrete state-action environments. \citet{liu2023benchmarking} proposes the Variational Inverse Constrained Reinforcement Learning (VICRL) method, which models the posterior distribution of the constraints, and is applicable to (high-dimensional) continuous state-action environments. Though, some prior methods use approximate Bayesian techniques to return a distribution over constraints, they do not have a way to identify the least constraining constraint that satisfies a degree of confidence. Also, prior methods do not have a way of determining if the expert trajectories available are sufficient to learn constraints with the desired level of confidence. Our CA-ICRL method addresses these two limitations of prior work. An alternative approach to ICRL is to adapt methods from the inverse reinforcement learning (IRL literature). For example, the popular IRL algorithm, Generative Adversarial Imitation Learning (GAIL) can be adapted to the ICRL setting by directly modifying the reward function to provide large punishments for violating the constraints \cite{liu2023benchmarking}. However in such methods, after learning punishments or constraints that are then re-used in new environments, the type of generalization obtained is different.  Learned punishments do not necessarily disable a target behaviour in new environments, depending on the transition dynamics and other rewards of that environment.  In contrast, learned constraints ensure that a target behaviour will remain disabled in other environments.

\section{Implementation and Environment Details}\label{appendix:implementation}

We provide additional details regarding the pseudocode of CA-ICRL in Algorithm 1.  Algorithm 1 iterates over 3 steps:
\begin{enumerate}
\item Forward Control: $\pi^* = \arg \max_\pi E_{P_\pi}[\bar{r}(\tau)]+\beta H(P_\pi)$ such that $E_{P_C}[\phi^*(\tau)]\le\epsilon$.  This constrained optimization can be solved by PPO-Lagrange~\cite{ray2019benchmarking} or PPO-Penalty~\cite{gaurav2023learning}.  In the experiments we use PPO-Lagrange.
\item Update weights of constraint distribution: $w \gets w + \nabla_w  [\sum_{\tau\in D} \beta\bar{r}(\tau) + \log quantile_{beta(\cdot|\alpha_w(D,\tau)}(1-\lambda) - \log Z(w)]$.  We provide more details about the gradient computation.  First note that the quantile function consists of the inverse cumulative distribution function (cdf).
\begin{equation}
quantile_{beta(\cdot|\alpha_w(D,\tau))}(1-\lambda) = cdf^{-1}_{beta(\cdot|\alpha_w(D,\tau)}(1-\lambda)
\end{equation}
Note that the cdf of the beta distribution is the regularized incomplete beta function. Since the inverse cdf of the beta distribution does not have a closed form, we approximate it numerically by a lookup table with inputs ranging from 0 to 1 in increments of 0.001.  To differentiate through this numerically approximated quantile function, we approximate the gradient by a finite forward difference (with $h=0.002$).
\begin{equation}
\nabla_w quantile_{beta(\cdot|\alpha_{w}(D,\tau))}(1-\lambda) \approx \frac{quantile_{beta(\cdot|\alpha_{w+h}(D,\tau))}(1-\lambda) - quantile_{beta(\cdot|\alpha_{w}(D,\tau))}(1-\lambda)}{h}
\end{equation}
where $h = 0.002$. 

To compute the $\nabla_w logZ(w)$ we use importance sampling as done in \citet{malik2021inverse}. Particularly, \citet{malik2021inverse} show that 

\begin{equation}\label{eq:importance}
    \nabla_w log Z(w) \approx \frac{1}{M} \sum_{j=1}^{M} \nabla_w \log \phi_w^* (\tau^j)
\end{equation}

\citet{malik2021inverse} use samples from an older policy $\pi_{\overline{w}}$, where $\overline{w}$ denotes the weights of $\phi$ at a previous iteration. To correct for the bias in importance sampling a weight $\nu$ is added given by 

\begin{equation}
    \nu(\tau) =  \frac{\phi_w^*(\tau)}{\phi_{\overline{w}}^*(\tau)}
\end{equation}

The gradient can then be further approximated by 

\begin{equation}
    \nabla_w log Z(w) \approx \frac{1}{M} \sum_{j=1}^M  \nu(\tau^j) \nabla \log \phi_w^*(\tau)
\end{equation}

where the trajectories $\{\tau^j\}_{j=1}^M$ are samples from $\pi_{\overline{w}}$. 

\item The third step involves computing $\phi^*(\tau)$ based on the confidence $\lambda$. We use the expression $\quad \phi^*(\tau) \gets quantile_{beta(\cdot|\alpha_w(D,\tau))}(1-\lambda)$. To compute the expectation over $\phi^*(\tau)$ in the forward control (step 1), we use a sample based approximation for the expectation using 50 trajectories.

\end{enumerate}

\begin{figure}
    \centering
    \includegraphics[width=.44\textwidth, height = 3cm]{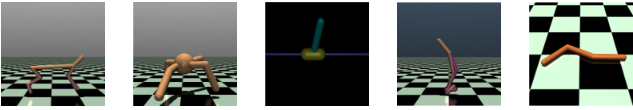}
    \caption{An illustration of the MuJoCo environments. These are the Half-Cheetah, Ant, Inverted Pendulum, Walker and Swimmer environments (from left to right).}
    \label{fig:MuJoCo environment}
\end{figure}

\begin{table*}
    \centering
    \begin{tabular}{|c|c|c|c|}
    \hline
         Name & Obs. Dim. & Act. Dim. 
         & Constraints \\
         \hline \hline
         Blocked Half-cheetah & 18 & 6 & X-Coordinate $\geq$ -3  \\
         \hline
         Blocked Ant & 113 & 8 & X-Coordinate $\geq$ -3   \\
         \hline
         Biased Pendulumn  & 4 & 1 & X-Coordinate $\geq$ -0.015   \\
         \hline
         Blocked Walker  & 18 & 6 & X-Coordinate $\geq$ -3   \\
         \hline
         Blocked Swimmer  & 10 & 2 &  X-Coordinate $\leq$ 0.5  \\
         \hline
    \end{tabular}
    \caption{Descriptions of the MuJoCo environments}
    \label{tab:mujoco}
\end{table*}

We use a set of MuJoCo environments, with settings very similar to that of \citet{liu2023benchmarking}. Table~\ref{tab:mujoco} tabulates the settings in the different environments. The constraints used for these experiments are the same as that in \citet{liu2023benchmarking}. A detailed analysis for the particular choices of these constraints can be found in \citet{liu2023benchmarking}.

The first domain is the Blocked Half-Cheetah domain shown in Figure~\ref{fig:MuJoCo environment}, where the agent is a robot on two legs with 9 links and 8 joints connecting the legs. The actions apply a torque on the joints that makes the agent move forward as fast as possible. The agent gets a positive reward proportional to the distance moved forward and a negative reward proportional to the distance moved backward. Each episode in this game has a maximum of 1000 time steps and the constraint has the robot stay between -3 and $\infty$ ($X \leq -3$). The second domain is the Blocked Ant domain where the agent is a robot with four legs, one torso and each leg has two links (a total of eight hinges). The actions apply torques on the eight hinges making the agent move forward. This domain has a maximum of 1000 time steps for each episode and a constraint of $X \leq -3$ similar to the Blocked Half-Cheetah environment. The agent gets a forward reward and a healthy reward corresponding to the distance it has covered in the forward direction and length of time it stays within the bounds of the environment. The third domain is the Biased Pendulum domain that tries to balance a pole on a cart. Each episode has a maximum of 100 time steps and an episode terminates either when the maximum number of time steps has been reached or when the pole falls down. The agent gets a reward of 0.1 for each time step where the agent remains in $X \geq 0$ and a reward of 1 if the agent has $X \leq -0.01$. There is a monotonic increase in reward between 0.1 to 1 when the $X$ coordinate is within the range $-0.01 < X < 0$. The constraint is, $X \leq -0.015$ which prevents the agent from taking large moves to the left. However, the reward is higher for left moves, hence the agent has to resist the temptation of the reward to respect the constraint bounds. The fourth domain is the Blocked Walker domain which is a two legged agent with two dimensions consisting of legs, thighs and feet, all of which need to coordinate to make the agent move forward. The action applies torque to all of the hinges connecting the different parts. Each episode can have a maximum of 1000 time steps and the game ends when the robot falls after losing balance. The constraint prevents the robot from moving too far backwards ($X \leq -3$) same as the Blocked Half-Cheetah and the Blocked Ant domains. The last domain is the Blocked Swimmer domain where the agent tries to swim (or walk) by applying torque to the rotor that connects two links to form a linear chain. This agent is in a virtual ``two-dimensional pool'' and tries to move as fast as possible towards the forward direction. Similar to previous domains, each episode in this domain has a maximum of 1000 time steps. The agent is rewarded for moving forward proportional to the distance it covers (and punished for moving backward proportional to the distance moved backwards) and is penalized for taking actions that are too large. The constraint is $X \geq 0.5$, which prevents the robot from moving too far ahead, since it is easier to move ahead than move back in this domain, which is opposite to the Blocked Half-Cheetah and the Blocked Ant domains where it is easier for the agent to move back than forward. Following~\citet{liu2023benchmarking}, we are most interested in studying the performance of ICRL methods in stochastic environments with added noise in the transition functions. Hence, for all environments, at each step we have an added noise such that $p(s_{t+1}|s_t, a_t) = f(s_t,a_t) + \eta$, where $\eta \sim \mathcal{N}(\mu, \sigma)$, with $\mu = 0$ and $\sigma = 0.2$.

\begin{table*}
    \centering
    \begin{tabular}{|c|c|c|c|}
    \hline
         Name & Obs. Dim. & Act. Dim. 
         & Constraints \\
         \hline \hline
         HighD Distance Constraint & 76 & 2 & Car Distance $\geq$ 20 m \\
         \hline
         HighD Velocity Constraint & 76 & 2 & Car Velocity $\leq$ 40 m/s   \\
         \hline
    \end{tabular}
    \caption{Descriptions of the realistic environments}
    \label{tab:realisticenvironments}
\end{table*}

\begin{figure}
    \centering
    \includegraphics[width=.44\textwidth, height = 3cm]{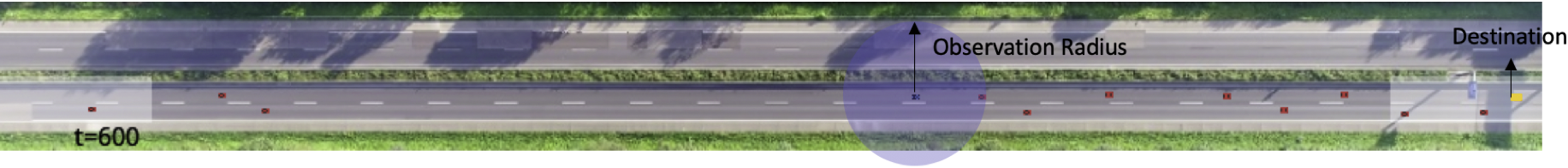}
    \caption{HighD Driving Environment. Figure from \citet{liu2023benchmarking}.}
    \label{fig:HighDEnvironment}
\end{figure}

The real-world driving environments use the same setting as \citet{liu2023benchmarking}. The environment is shown in Figure~\ref{fig:HighDEnvironment} and the constraints for this environment are shown in Table~\ref{tab:realisticenvironments}. In Figure~\ref{fig:HighDEnvironment} the ego car is in blue and the other cars are in red. The ego car is being controlled by the agent, which tries to drive it as efficiently as possible without violating constraints.

The hyperparameters are largely the same as those used by prior works~\cite{liu2023benchmarking, Baert2023Maximum}. These parameters are same for all the ICRL algorthms. For the virutal environments, PPO-Lag batch size is 64, hidden layer size is 64 and the number of hidden layers for policy, value and cost networks is 3. For the HighD driving environments, the batch size of the constraint model is 1000, the hidden layer size is 64 and the number of hidden layers for policy, value and cost networks is 3. Regarding CA-ICRL, we use transformers to implement the encoder blocks given in Figure~4. CA-ICRL contains 2 heads with 4 hidden layers. The value for $\beta = 0.02$ throughout. For all the experiments, we repeat 50 times for training (random seeds 1 -- 50) and 50 times for execution (random seeds 51 -- 100) and plot the averages and standard deviations of performances.

\section{Computational Complexity}\label{appendix:computationalcomplexity}

If $M$ is the total number of training episodes of length $L$, $N$ is the total number of expert trajectories, $K$ is the total number of iterations within an episodic step, and $X$ is the total number of policy trajectories used for constraint adjustment at each time step, the time complexity of the CA-ICRL algorithm can be given as $O(KLMNX)$. This is because each trajectory from the policy is compared against every expert trajectory to compute a similarity score in CA-ICRL. This happens for every iteration in every step in every episode of training.

\section{Computational Infrastructure and Wall Clock Times}\label{appendix:wallclocktimes}

All the training for the experiments were conducted on a virtual machine having 2 Nvidia A100 GPUs with a GPU memory of 40 GB. The CPUs use the AMD EPYC processors with a memory of 125 GB. The MuJoCo experiments took an average of 80 hours of wall clock time to complete and the experiments on the realistic driving environments took an average of 95 hours wall clock time to complete.

\section{Expected Calibration Error}\label{appendix:ECE}

\begin{figure*}
\centering
\begin{subfigure}{.18\textwidth}
  \centering
\includegraphics[width=3cm, height = 3cm]{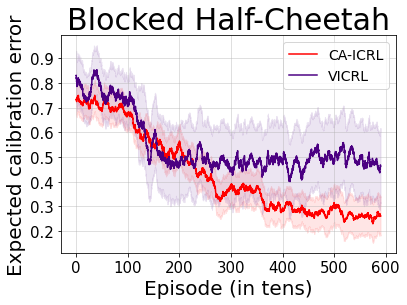}

\end{subfigure}%
\begin{subfigure}{.18\textwidth}
  \centering
  \includegraphics[width=3cm, height = 3cm]{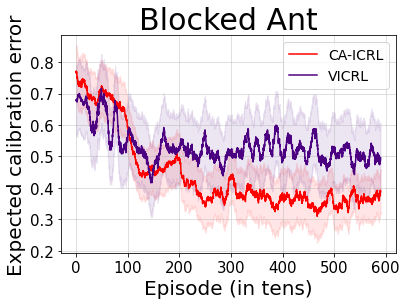}

\end{subfigure}
\begin{subfigure}{.18\textwidth}
  \centering
  \includegraphics[width=3cm, height = 3cm]{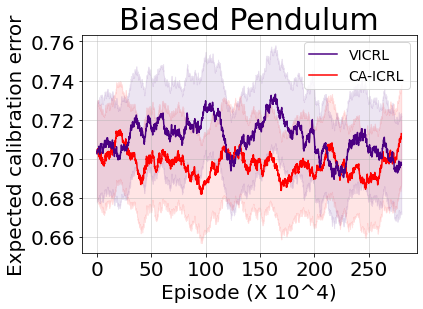}

\end{subfigure}
\begin{subfigure}{.18\textwidth}
  \centering
  \includegraphics[width=3cm, height = 3cm]{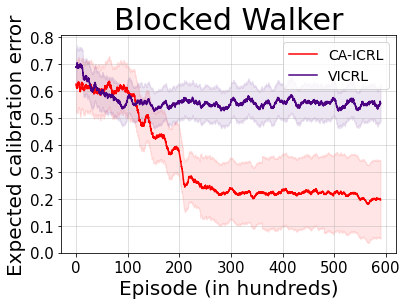}

\end{subfigure}
\begin{subfigure}{.18\textwidth}
  \centering
  \includegraphics[width=3cm, height = 3cm]{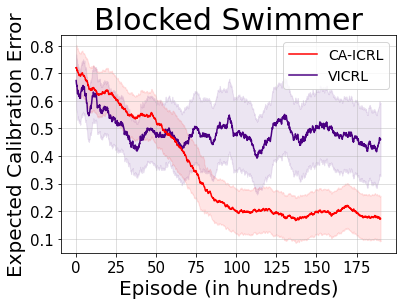}
\end{subfigure}

\caption{Expected Calibration Error in different MuJoCo environments during training (lower is better)}
\label{fig:ece}
\end{figure*}

In this section, we plot the expected calibration error (ECE) of CA-ICRL and VICRL in the different MuJoCo environments. These experiments use 150 expert trajectories for each of the environments, with a confidence of 70\% for CA-ICRL. The ECE pertains to the constraint adjustment network in both the methods, which returns a probability with which a given trajectory is feasible. To compute the ECE, we simulate a set of 500 trajectories, of which 250 are feasible and 250 are infeasible (ground truth label). We split the data into 5 equally spaced bins. If the probability of a trajectory is below 0.5, it is assigned the label of 0 (not feasible) and if the probability is above 0.5, it is assigned the label of 1 (feasible). We use the formula $ECE = \sum_{m-1}^M \frac{|B_m|}{n} |acc(B_m) - conf(B_m)|$ to compute the ECE \cite{ece}. We compute the ECE at every episode during training. The plots are shown in Figure~\ref{fig:ece}, where CA-ICRL provides a better ECE as compared to VICRL in four environments and a similar performance as compared to VICRL in Biased Pendulum. Since CA-ICRL learns a similarity score for each trajectory as compared to each of the expert trajectories (with direct comparisons to each expert trajectory), it provides a better estimate of the feasibility of a given trajectory. Comparatively, the VICRL method learns a distribution, but includes unbounded approximations to the posterior, which leads to a poor ECE (indirect comparisons to expert trajectories).

\section{Additional Experimental Results}\label{appendix:additionalresults}

\begin{figure*}
\centering
\begin{subfigure}{.33\textwidth}
  \centering
\includegraphics[width=5cm, height = 5cm]{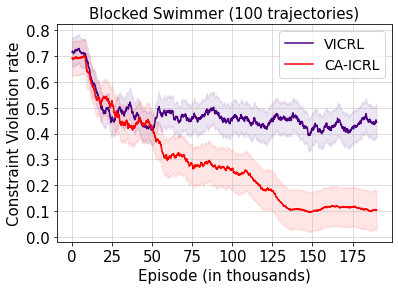}

\end{subfigure}%
\begin{subfigure}{.33\textwidth}
  \centering
  \includegraphics[width=5cm, height = 5cm]{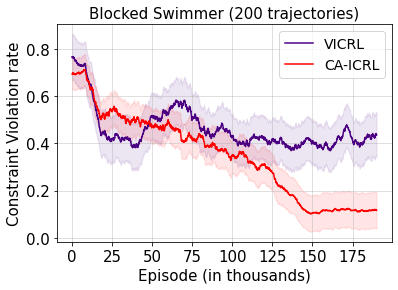}
 
\end{subfigure}
\begin{subfigure}{.33\textwidth}
  \centering
  \includegraphics[width=5cm, height = 5cm]{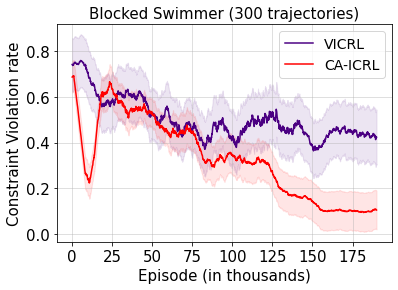} 
  
\end{subfigure}

\caption{Comparison of constraint violation rate between VICRL and CA-ICRL in the Blocked Swimmer Environment for different numbers of expert trajectories}
\label{fig:blockedswimmerdifferenttrajectoriesconstraintviolationrate}
\end{figure*}

\begin{figure*}
\centering
\begin{subfigure}{.33\textwidth}
  \centering
\includegraphics[width=5cm, height = 5cm]{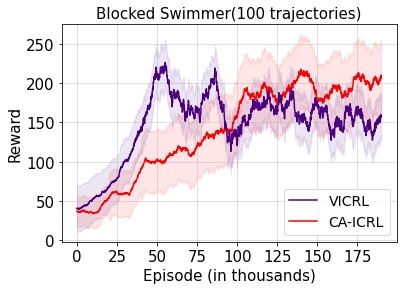}

\end{subfigure}%
\begin{subfigure}{.33\textwidth}
  \centering
  \includegraphics[width=5cm, height = 5cm]{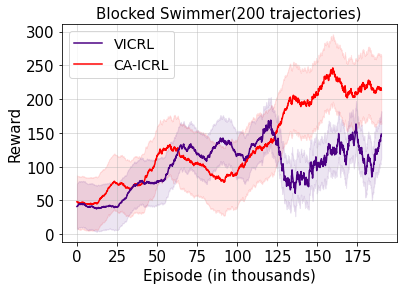}
 
\end{subfigure}
\begin{subfigure}{.33\textwidth}
  \centering
  \includegraphics[width=5cm, height = 5cm]{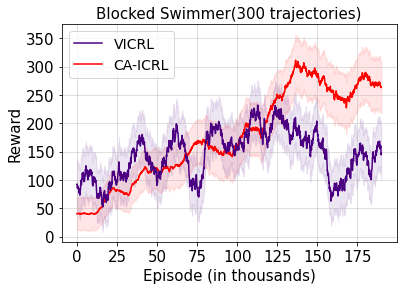} 
  
\end{subfigure}

\caption{Comparison of rewards earned between VICRL and CA-ICRL in the Blocked Swimmer Environment for different numbers of expert trajectories}
\label{fig:blockedswimmerdifferenttrajectoriesrewards}
\end{figure*}

\begin{figure*}
\centering
\begin{subfigure}{.33\textwidth}
  \centering
\includegraphics[width=5cm, height = 5cm]{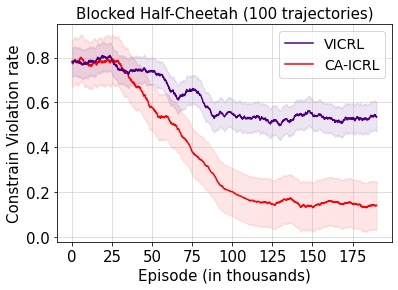}

\end{subfigure}%
\begin{subfigure}{.33\textwidth}
  \centering
  \includegraphics[width=5cm, height = 5cm]{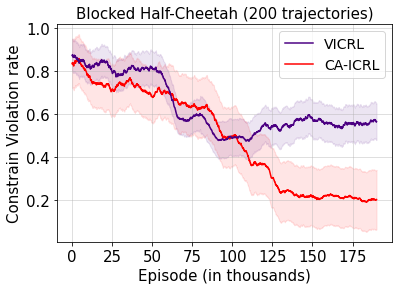}
 
\end{subfigure}
\begin{subfigure}{.33\textwidth}
  \centering
  \includegraphics[width=5cm, height = 5cm]{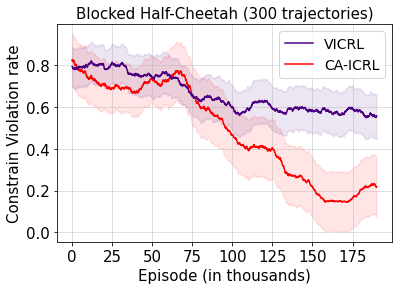} 
  
\end{subfigure}

\caption{Comparison of constraint violation rate between VICRL and CA-ICRL in the Blocked Half-Cheetah Environment for different numbers of expert trajectories}
\label{fig:blockedhalfcheetahdifferenttrajectoriesconstraintviolationrate}
\end{figure*}

\begin{figure*}
\centering
\begin{subfigure}{.33\textwidth}
  \centering
\includegraphics[width=5cm, height = 5cm]{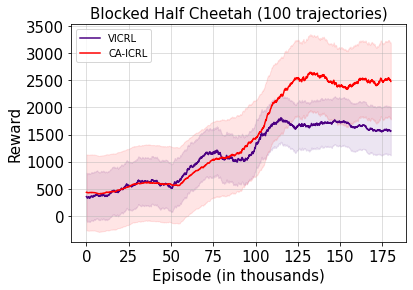}

\end{subfigure}%
\begin{subfigure}{.33\textwidth}
  \centering
  \includegraphics[width=5cm, height = 5cm]{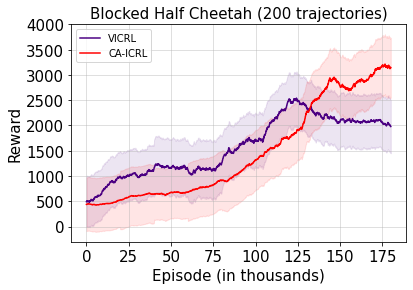}
 
\end{subfigure}
\begin{subfigure}{.33\textwidth}
  \centering
  \includegraphics[width=5cm, height = 5cm]{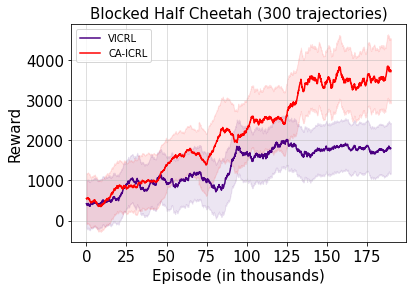} 
  
\end{subfigure}

\caption{Comparison of rewards earned between VICRL and CA-ICRL in the Blocked Half-Cheetah Environment for different numbers of expert trajectories}
\label{fig:blockedhalfcheetahdifferenttrajectoriesrewards}
\end{figure*}

\begin{figure*}
\centering
\begin{subfigure}{.33\textwidth}
  \centering
\includegraphics[width=5cm, height = 5cm]{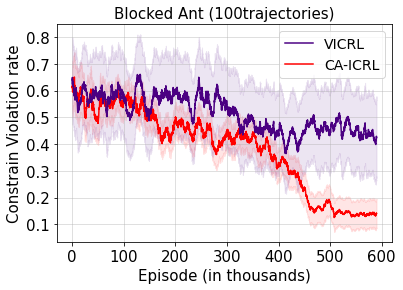}

\end{subfigure}%
\begin{subfigure}{.33\textwidth}
  \centering
  \includegraphics[width=5cm, height = 5cm]{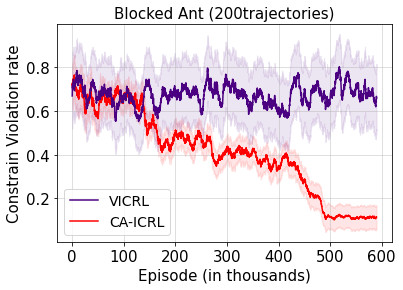}
 
\end{subfigure}
\begin{subfigure}{.33\textwidth}
  \centering
  \includegraphics[width=5cm, height = 5cm]{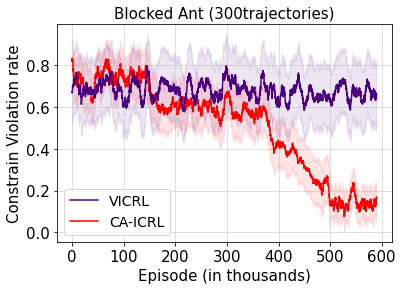} 
  
\end{subfigure}

\caption{Comparison of constraint violation rate between VICRL and CA-ICRL in the Blocked Ant Environment for different numbers of expert trajectories}
\label{fig:blockedantdifferenttrajectoriesconstraintviolationrate}
\end{figure*}

\begin{figure*}
\centering
\begin{subfigure}{.33\textwidth}
  \centering
\includegraphics[width=5cm, height = 5cm]{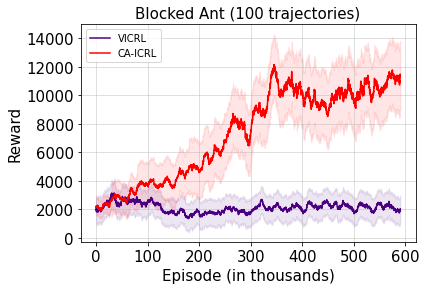}

\end{subfigure}%
\begin{subfigure}{.33\textwidth}
  \centering
  \includegraphics[width=5cm, height = 5cm]{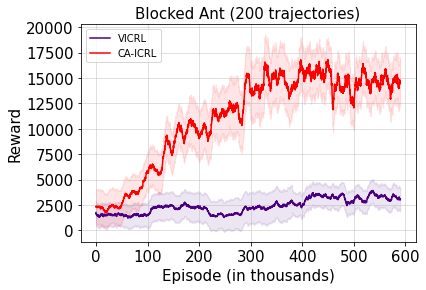}
 
\end{subfigure}
\begin{subfigure}{.33\textwidth}
  \centering
  \includegraphics[width=5cm, height = 5cm]{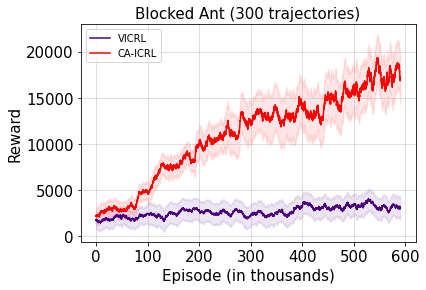} 
  
\end{subfigure}

\caption{Comparison of rewards earned between VICRL and CA-ICRL in the Blocked Ant Environment for different numbers of expert trajectories}
\label{fig:blockedantdifferenttrajectoriesrewards}
\end{figure*}

\begin{figure*}
\centering
\begin{subfigure}{.30\textwidth}
  \centering
\includegraphics[width=5cm, height = 5cm]{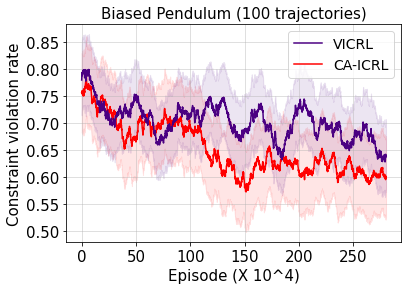}

\end{subfigure}%
\begin{subfigure}{.30\textwidth}
  \centering
  \includegraphics[width=5cm, height = 5cm]{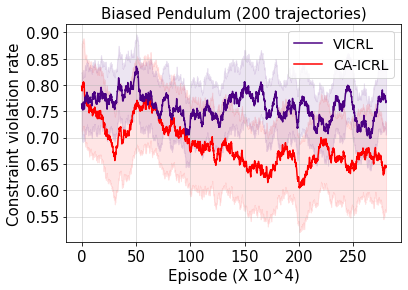}
 
\end{subfigure}
\begin{subfigure}{.30\textwidth}
  \centering
  \includegraphics[width=5cm, height = 5cm]{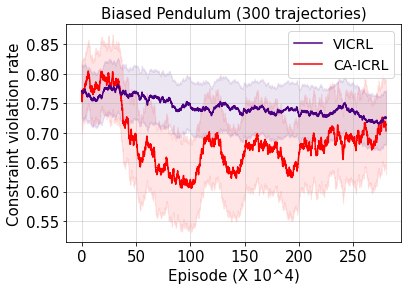} 
  
\end{subfigure}

\caption{Comparison of constraint violation rate between VICRL and CA-ICRL in the Biased Pendulum Environment for different numbers of expert trajectories}
\label{fig:Biasedpendulumdifferenttrajectoriesconstraintviolationrate}
\end{figure*}

\begin{figure*}
\centering
\begin{subfigure}{.33\textwidth}
  \centering
\includegraphics[width=5cm, height = 5cm]{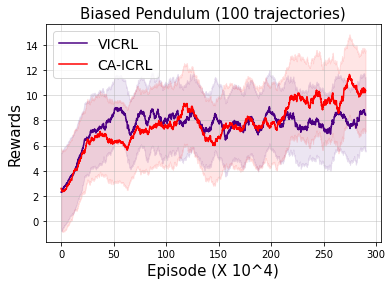}

\end{subfigure}%
\begin{subfigure}{.33\textwidth}
  \centering
  \includegraphics[width=5cm, height = 5cm]{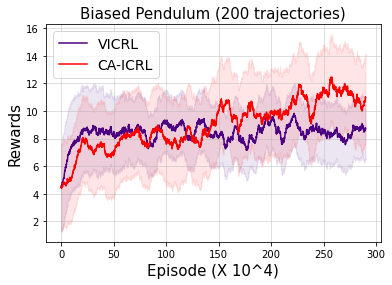}
 
\end{subfigure}
\begin{subfigure}{.33\textwidth}
  \centering
  \includegraphics[width=5cm, height = 5cm]{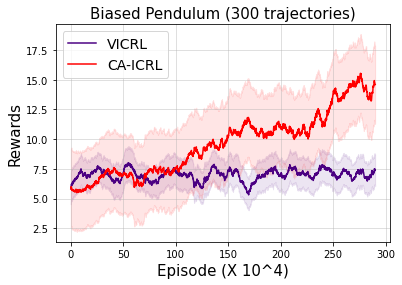} 
  
\end{subfigure}

\caption{Comparison of rewards earned between VICRL and CA-ICRL in the Biased Pendulum Environment for different numbers of expert trajectories}
\label{fig:biasedpendulumdifferenttrajectoriesrewards}
\end{figure*}

In Figures~\ref{fig:blockedswimmerdifferenttrajectoriesconstraintviolationrate} -- \ref{fig:biasedpendulumdifferenttrajectoriesrewards} we show additional results comparing the performances of CA-ICRL and VICRL while the number of expert trajectories is varied (100, 200 and 300 trajectories). These experiments use four MuJoCo environments (Blocked Swimmer, Blocked Half-Cheetah, Blocked Ant, Biased Pendulum), the experiments in Blocked Walker can be found in the experimental section of the main paper. These experiments correspond to our second objective, where CA-ICRL can be used by practitioners to know if the number of expert trajectories are sufficient for a given task with a desired level of confidence and performance (through rewards). As noted in the main paper, in most of the experiments in this section, we note that CA-ICRL performs better than VICRL (both in terms of the constraint violation rate and the rewards obtained). Also, CA-ICRL obtains higher rewards as the number of expert trajectories increase and consistently learns a policy that has a constraint violation rate lower than required. The only exception to this observation is in the Biased Pendulum experiment (Figure~\ref{fig:Biasedpendulumdifferenttrajectoriesconstraintviolationrate} and Figure~\ref{fig:biasedpendulumdifferenttrajectoriesrewards}) where the constraint violation rate remains quite high. As noted previously, Biased Pendulum is a very hard domain where all ICRL methods struggle to learn good policies that provide high rewards and have low constraint violation rates. Nonetheless, we note that CA-ICRL still outperforms VICRL in the Biased Pendulum experiments across all the three scenarios (100, 200 and 300 expert trajectories).

Table~\ref{table:statsigtraining} and Table~\ref{table:statsigexecution} provides the $p$-values of an unpaired 2-sided t-test for statistical significance in the training and execution experiments respectively. The $p$-values are for the comparisons of performances each algorithm with that CA-ICRL in each of the execution experiments conducted in experimental section of the main paper. From the $p$-values we see most of our inferences showing the superiority of CA-ICRL are statistically significant (except a few results in Biased Pendulum).

\begin{table*}[htbp]
\centering
\resizebox{0.9\textwidth}{!}{
\begin{tabular}{c|cccccccc}
\toprule
 & & \begin{tabular}[c]{@{}c@{}}Blocked Half-\\ Cheetah\end{tabular} & \begin{tabular}[c]{@{}c@{}}Blocked\\ Ant\end{tabular} & \begin{tabular}[c]{@{}c@{}}Biased\\ Pendulum\end{tabular} & \begin{tabular}[c]{@{}c@{}}Blocked\\ Walker\end{tabular} & \begin{tabular}[c]{@{}c@{}}Blocked\\ Swimmer\end{tabular} & \begin{tabular}[c]{@{}c@{}}HighD \\ Distance\end{tabular} & \begin{tabular}[c]{@{}c@{}}HighD \\ Velocity\end{tabular} \\ \hline
\multirow{4}{*}{\begin{tabular}[c]{@{}c@{}} Rewards\end{tabular}} & GACL & 0.016  & 0.035  & \textbf{0.063} & 0.024 & 0.001 & 0.004 & 0.036 \\
& BC2L & 0.018  & 0.029  & \textbf{0.071} & 0.035 & 0.001 & 0.004 & 0.025 \\
& ICRL & 0.001  & 0.001  & 0.033 & 0.001 & 0.027 & 0.004 & 0.023 \\
& VICRL  & 0.001  & 0.001  & 0.001 & 0.001 & 0.001 & 0.001 & 0.013\\\hline
\multirow{4}{*}{\begin{tabular}[c]{@{}c@{}}Constraint\\Violation\\ Rate\end{tabular}} & GACL & 0.001  & 0.001  & 0.022 & 0.001 & 0.001 & 0.001 & 0.001 \\
& BC2L & 0.001  & 0.001  & 0.024 & 0.001 & 0.001 & 0.001 & 0.001 \\
& ICRL & 0.001  & 0.001  & 0.017 & 0.001 & 0.001 & 0.001 & 0.001 \\
& VICRL  & 0.001  & 0.019  & 0.001 & 0.001 & 0.014 & 0.005 & 0.001 \\\bottomrule
\end{tabular}
\vspace{-0.05in}
}
\vspace{-0.05in}
\caption{Statistical Significance - We report the $p$-values from an unpaired 2-sided t-test with comparisons to CA-ICRL for the training experiments, conducted at the last episode of training. The values are rounded to the third decimal. We consider $p$ < 0.05 as statistically significant. From the table, most comparisons are statistically significant differences (except the ones in bold). }\label{table:statsigtraining}
\vspace{-0.05in}
\end{table*}

\begin{table*}[htbp]
\centering
\resizebox{0.9\textwidth}{!}{
\begin{tabular}{c|cccccccc}
\toprule
 & & \begin{tabular}[c]{@{}c@{}}Blocked Half-\\ Cheetah\end{tabular} & \begin{tabular}[c]{@{}c@{}}Blocked\\ Ant\end{tabular} & \begin{tabular}[c]{@{}c@{}}Biased\\ Pendulum\end{tabular} & \begin{tabular}[c]{@{}c@{}}Blocked\\ Walker\end{tabular} & \begin{tabular}[c]{@{}c@{}}Blocked\\ Swimmer\end{tabular} & \begin{tabular}[c]{@{}c@{}}HighD \\ Distance\end{tabular} & \begin{tabular}[c]{@{}c@{}}HighD \\ Velocity\end{tabular} \\ \hline
\multirow{4}{*}{\begin{tabular}[c]{@{}c@{}} Rewards\end{tabular}} & GACL & 0.043  & 0.039  & \textbf{0.081} & 0.001 & 0.001 & 0.015 & 0.001 \\
& BC2L & \textbf{0.056}  & 0.045  & \textbf{0.086} & 0.035 & 0.001 & 0.031 & 0.001 \\
& ICRL & 0.001  & 0.001  & \textbf{0.053} & 0.001 & 0.027 & 0.048 & 0.043 \\
& VICRL  & 0.001  & 0.001  & 0.001 & 0.001 & 0.001 & 0.026 & 0.039\\\hline
\multirow{4}{*}{\begin{tabular}[c]{@{}c@{}}Constraint\\Violation\\ Rate\end{tabular}} & GACL & 0.001  & 0.001  & 0.007 & 0.001 & 0.001 & 0.001 & 0.001 \\
& BC2L & 0.001  & 0.008  & 0.031 & 0.001 & 0.001 & 0.001 & 0.001 \\
& ICRL & 0.003  & 0.001  & \textbf{0.068} & 0.006 & 0.001 & 0.007 & 0.001 \\
& VICRL  & 0.001  & 0.019  & \textbf{0.084} & 0.033 & 0.014 & 0.004 & 0.002 \\\bottomrule
\end{tabular}
\vspace{-0.05in}
}
\vspace{-0.05in}
\caption{Statistical Significance - We report the $p$-values from an unpaired 2-sided t-test with comparisons to CA-ICRL for the testing experiments. The values are rounded to the third decimal. We consider $p < 0.05$ as statistically significant differences. From the table, most comparisons are statistically significant differences (except the ones in bold). }\label{table:statsigexecution}
\vspace{-0.05in}
\end{table*}


\section{Experiments With Thousands of Trajectories}\label{appendix:thousandsoftrajectories}

\begin{figure*}
\centering
\begin{subfigure}{.40\textwidth}
\includegraphics[width=\textwidth, height = 5cm]{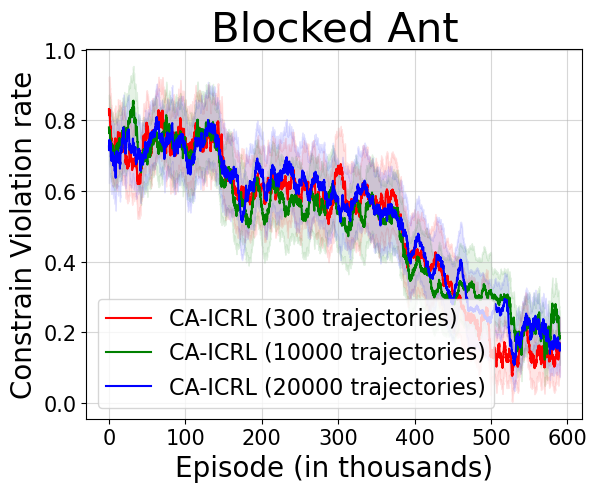}
\end{subfigure}%
\quad \quad \quad 
\begin{subfigure}{.40\textwidth}
  \includegraphics[width=\textwidth, height=5cm]{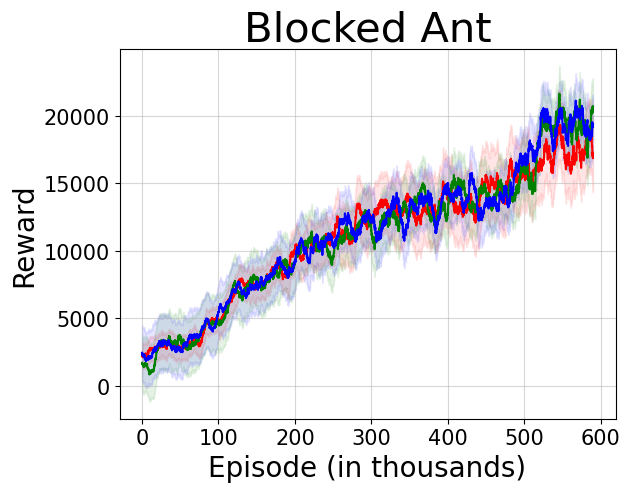}
 \end{subfigure}
\\
\begin{subfigure}{.40\textwidth}
  \includegraphics[width=\textwidth, height=5cm]{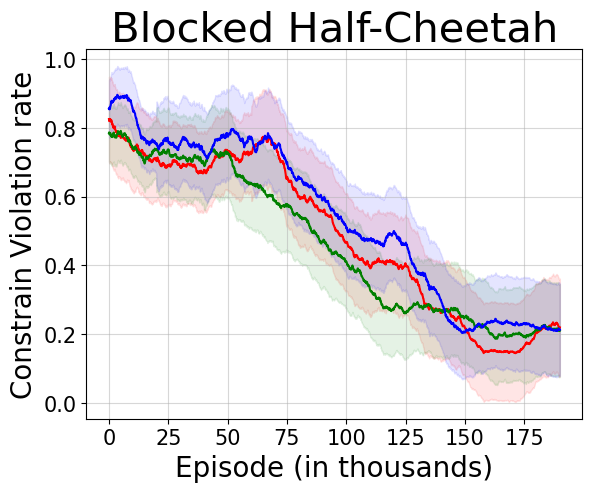} 
  \end{subfigure}
  \quad \quad \quad
\begin{subfigure}{.40\textwidth}
  \includegraphics[width=\textwidth, height=5cm]{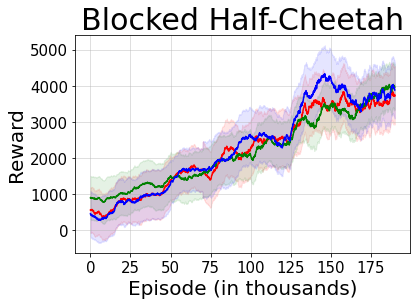} 
  \end{subfigure}

\caption{CA-ICRL with thousands of trajectories. All figures share the same legend.}
\label{fig:caicrlthousanddifferenttrajectories}
  \end{figure*}

In the experiments in Section~\ref{sec:experiments} we restrict ourselves to hundreds of trajectories since that was close to sufficient in these environments. Figure~\ref{fig:caicrlthousanddifferenttrajectories} shows results with up-to 20,000 trajectories (with performances only showing a small improvement over 300 trajectories).

\section{Comparisons With ICL~\cite{gaurav2023learning}}\label{appendix:ICLcomparisons}

\begin{figure*}
\centering
\begin{subfigure}{.40\textwidth}
\includegraphics[width=\textwidth, height = 5cm]{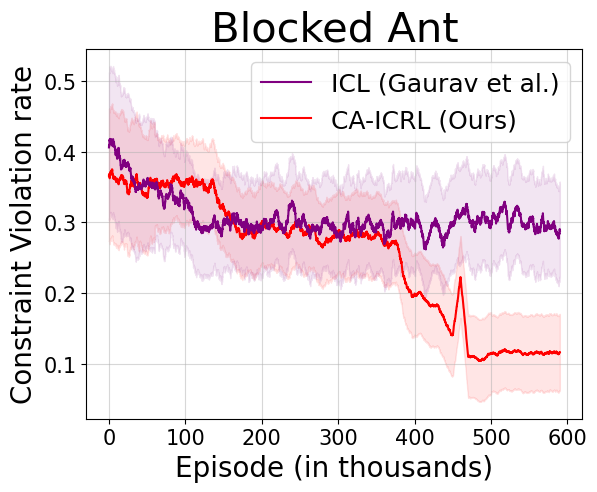}
\end{subfigure}%
\quad \quad \quad 
\begin{subfigure}{.40\textwidth}
  \includegraphics[width=\textwidth, height=5cm]{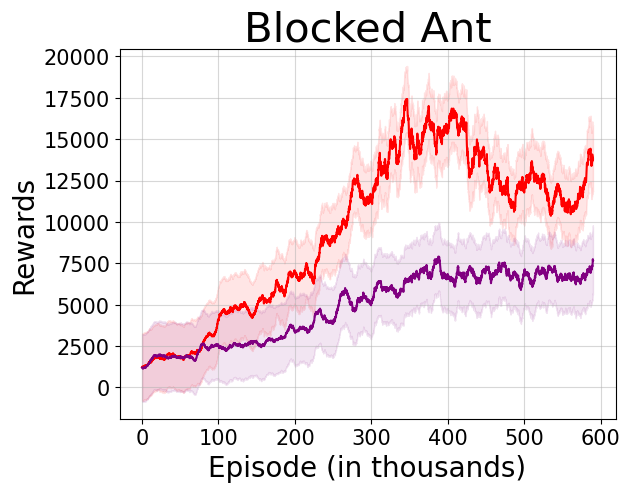}
 \end{subfigure}
\\
\begin{subfigure}{.40\textwidth}
  \includegraphics[width=\textwidth, height=5cm]{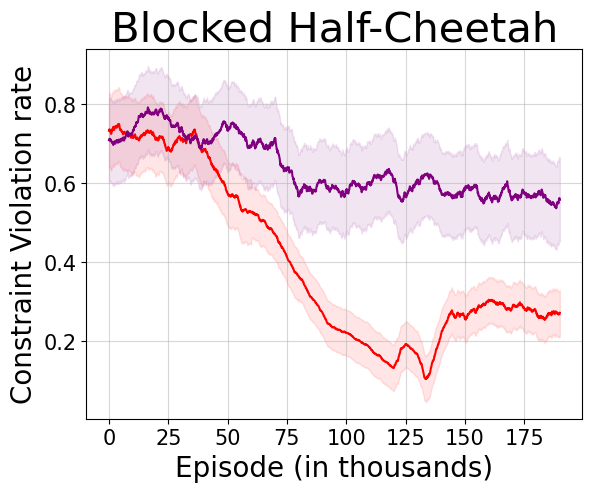} 
  \end{subfigure}
  \quad \quad \quad
\begin{subfigure}{.40\textwidth}
  \includegraphics[width=\textwidth, height=5cm]{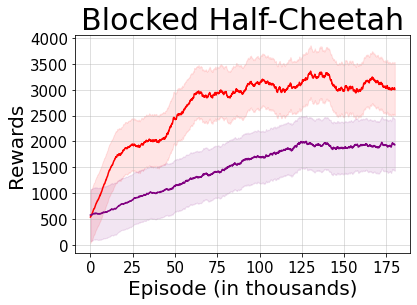} 
  \end{subfigure}

\caption{Comparison of CA-ICRL with ICL~\cite{gaurav2023learning}. All figures share the same legend.}
\label{fig:comparisonwithICL}
  \end{figure*}

In the experiments in Section~\ref{sec:experiments}, we have not considered comparisons to \citet{gaurav2023learning} since this method neither learns confidence nor a distribution over constraints (see Table~\ref{tab:ICRLalgorithms}). However, for completeness we run some comparisons to ICL as well in two MuJoCo environments (Blocked Ant and Blocked Half-Cheetah). The results are in Figure~\ref{fig:comparisonwithICL} which demonstrate that CA-ICRL shows consistently better performances as compared to ICL across the different two different metrics (i.e., constraint violation rate and rewards obtained).

\end{document}